\def\ARXIVVERSION{1}
\def\ARXIVAUTHORS{Runlin Shi$^{1,2}$ \quad Bojian Yin$^{1,*}$ \quad Guoqi Li$^{1,*}$\\
$^1$Institute of Automation, Chinese Academy of Sciences, Beijing, China\\
$^2$School of Future Technology, University of Chinese Academy of Sciences, Beijing, China\\
$^*$Corresponding authors.}
\def\ARXIVPDFAUTHORS{Runlin Shi, Bojian Yin, Guoqi Li}

\documentclass{article} 
\usepackage{iclr2027_conference,times}

\ifdefined\ARXIVVERSION
\iclrfinalcopy
\fi

\usepackage{amsmath,amsfonts,bm}

\def\eqref#1{equation~\ref{#1}}

\def\1{\bm{1}}

\DeclareMathAlphabet{\mathsfit}{\encodingdefault}{\sfdefault}{m}{sl}
\SetMathAlphabet{\mathsfit}{bold}{\encodingdefault}{\sfdefault}{bx}{n}

\usepackage{hyperref}
\usepackage{url}
\usepackage{graphicx}
\usepackage{booktabs}
\usepackage{amssymb}
\usepackage{placeins}

\hypersetup{pdftitle={Two Principles for Autoregressive Language Modeling}}
\ifdefined\ARXIVVERSION
\hypersetup{pdfauthor={\ARXIVPDFAUTHORS}}
\fi

\title{Two Principles for\\Autoregressive Language Modeling}

\ifdefined\ARXIVVERSION
\author{\ARXIVAUTHORS}
\else
\author{}
\fi

\begin{document}

\ifdefined\ARXIVVERSION
\pagestyle{plain}
\fi

\maketitle

\begin{abstract}
Improving large language model (LLM) architectures requires principled guidance on functional organization. We seek such guidance in the functional differentiation implicitly learned by RoPE-based Transformers. Behavioral retrieval and local-streaming probes reveal head-level tendencies but not a complete functional taxonomy. We therefore propose two faithful intervention-based metrics: RoPE Frequency Importance Score (RFIS), which measures how each frequency contribution affects a head's complete attention distribution, and RoPE Positional Dependence (RPD), isolating the effect of rotary positional modulation. In Qwen3-series models and Llama3.1, these metrics reveal a complete mechanism-level taxonomy of retrieval and positional modeling, with a salient mid-low-frequency band separating their frequency dependencies. Heads tend to specialize but can exhibit both functions. Controlled Transformer experiments show that this Global Positional Band (GPBand) tracks the positional scale set by the training sequence length, suggesting that global positional dependence may contribute to zero-shot length-extrapolation failure. Together with the layer-specific distribution of functions, these findings motivate two design principles: \textbf{(1) Functional Organization:} retrieval and positional functions should be assigned separately and mixed within layers with layer-specific allocation; and \textbf{(2) Positional Dependence:} positional modeling should operate only locally, while global access should rely on position-independent retrieval. We test these principles through matched from-scratch pretraining and targeted architectural ablations at 380M parameters, and scale up to 1.4B-parameter. The results support the effectiveness of both principles and highlight the importance of mechanism choice in realizing each function. Together, these findings suggest a promising route toward architectural innovation grounded in the functional requirements of autoregressive language modeling.
\end{abstract}

\section{Introduction}
\label{sec:introduction}
Improving large language model (LLM) architectures calls for understanding which functions they should realize and how those functions should be organized. We seek an empirical basis in the functional differentiation implicitly learned by RoPE-based Transformers, asking whether making this differentiation explicit can improve autoregressive language modeling.

Prior work identifies retrieval heads that support long-range access~\citep{wuRetrievalHeadMechanistically2024,zhangQueryFocusedRetrievalHeads2025}. Subsequent methods exploit this specialization for post-training KV-cache optimization or checkpoint conversion~\citep{xiaoDuoAttentionEfficientLongContext2024,tanHydraHeadHeadLevelFunctional2026}. We seek to move from exploiting specialization in pretrained checkpoints to characterizing the complementary functions, their properties, and the principles for organizing them in architectures trained from scratch.

To establish a complete mechanism-level taxonomy, we begin with behavioral probes of head-level functional differentiation. We use QRscore~\citep{zhangQueryFocusedRetrievalHeads2025} to measure retrieval tendency and introduce LDscore to measure local-streaming tendency. The two probes reveal behavioral tendencies and their layer-specific distribution but do not yield a complete functional taxonomy.
We consequently turn to the underlying RoPE-frequency dependence and propose two faithful intervention-based metrics. \emph{RoPE Frequency Importance Score} (RFIS) measures how removing one frequency contribution changes a head's complete attention distribution, whereas \emph{RoPE Positional Dependence} (RPD) isolates whether that effect arises from rotary positional modulation. RFIS reveals contrasting frequency-dependence patterns associated with retrieval and positional modeling, and RPD verifies their positional interpretation. Heads with a predominant retrieval tendency emphasize frequencies below a salient mid-low-frequency band, whereas heads with stronger positional dependence emphasize the band and the frequencies above it. Individual heads can exhibit both patterns. Together, these results establish a complete mechanism-level taxonomy of retrieval and positional modeling in the RoPE-frequency dependence we analyze.

Controlled Transformers further show that this functional boundary follows the training-length positional scale. We therefore term it the \emph{Global Positional Band} (GPBand). The associated global positional fitting suggests a potential failure mode beyond the training length. Together with the depth-varying head distribution, these findings motivate two design principles: \textbf{(1) Functional Organization:} retrieval and positional functions should be assigned separately and mixed within layers with layer-specific allocation; and \textbf{(2) Positional Dependence:} positional modeling should operate only locally, while global access should rely on position-independent retrieval.

We test these principles through matched from-scratch pretraining and targeted architectural ablations at $380$M parameters, and scale up to $1.4$B-parameter models for further validation. The layer-wise hybrid Architecture partially implements the principles through functional separation across layers. The proposed Head-wise Hybrid Architecture (HwH) serves as an experimental framework: its reference design assigns NoPE Full Attention (FA) to global retrieval and Linear Attention (LA) to local positional modeling, while its variants test functional allocation and component choices. Comparisons with Transformer and pure-LA baselines, together with component and allocation ablations, support both principles and highlight the importance of mechanism choice in realizing each function.

Our contributions are summarized as follows:
\begin{itemize}
\item We propose RFIS and RPD, faithful and bounded intervention-based metrics for RoPE-frequency analysis, and use them to establish a complete mechanism-level taxonomy of retrieval and positional modeling in RoPE Transformers.
\item We characterize the training-length-related salient mid-low-frequency band in RoPE Transformers as the functional boundary of this taxonomy, term it GPBand, and identify global positional fitting as a potential cause of extrapolation failure.
\item We derive two architectural design principles and test them through matched from-scratch comparisons and targeted allocation and component ablations, providing evidence for both principles and the importance of mechanism choice.
\item We reveal RoPE Transformers as implicit functional hybrids and suggest a promising route toward architectural innovation grounded in the functional requirements of autoregressive language modeling.
\end{itemize}

\section{Preliminaries}
\label{sec:preliminaries}

\paragraph{Full attention and grouped-query attention.}
\label{subsec:softmax_attention}

For a sequence $X\in\mathbb R^{T\times d}$, FA~\citep{vaswaniAttentionAllYou2017} forms $Q=XW^Q$, $K=XW^K$, and $V=XW^V$. A head of dimension $d_h$ outputs
\begin{equation}
    O=\operatorname{softmax}\!\left(
        \frac{QK^\top}{\sqrt{d_h}}+M
    \right)V ,
    \label{eq:softmax_attention}
\end{equation}
where $M$ is the causal mask. FA requires $O(T^2)$ pairwise interactions, and its KV cache grows linearly with context. Grouped-query attention (GQA)~\citep{ainslieGQATrainingGeneralized2023} shares each of $G\leq H$ KV heads among a group of $H$ Query heads; let $g(h)$ map Query head $h$ to its KV head. The cases $G=H$ and $G=1$ recover MHA and MQA~\citep{shazeerFastTransformerDecoding2019}, respectively. We measure each Query head separately because it retains its own query projection and attention distribution.

\paragraph{Linear attention.}
\label{subsec:linear_attention}

LA replaces pairwise FA with a fixed-size recurrent state~\citep{katharopoulosTransformersAreRNNs2020}. A basic recurrence is
\begin{equation}
    S_t=S_{t-1}+v_tk_t^\top,
    \qquad
    o_t=S_tq_t,
    \label{eq:linear_attention_recurrent}
\end{equation}
where $S_t\in\mathbb R^{d_v\times d_k}$, giving linear sequence complexity and constant-size recurrent memory for fixed dimensions. We use Gated DeltaNet (GDN)~\citep{yangGatedDeltaNetworks2025}, which adds data-dependent decay and a delta-rule update:
\begin{equation}
    S_t =
    S_{t-1}\bigl(\alpha_t(I-\beta_t k_t k_t^\top)\bigr)
    + \beta_t v_t k_t^\top,
    \qquad
    o_t = S_t q_t ,
    \label{eq:gated_delta_rule}
\end{equation}
Here $\alpha_t,\beta_t\in(0,1)$ control retention and updating.

\paragraph{Rotary position embedding.}
\label{subsec:rope}

Rotary Position Embedding (RoPE) partitions each Query and Key head into $F=d_h/2$ two-dimensional groups and rotates group $r\in\{0,\ldots,F-1\}$ by frequency $\omega_r$~\citep{suRoFormerEnhancedTransformer2023}:
\begin{equation}
\begin{aligned}
    \omega_r&=\theta^{-2r/d_h},
    &R(\varphi)&=
    \begin{bmatrix}
        \cos\varphi&-\sin\varphi\\
        \sin\varphi& \cos\varphi
    \end{bmatrix},\\
    \tilde q_t^{(r)}&=R(t\omega_r)q_t^{(r)},
    &\tilde k_s^{(r)}&=R(s\omega_r)k_s^{(r)} .
\end{aligned}
    \label{eq:rope_frequency}
\end{equation}
Here $\theta$ is the RoPE base and $r=0$ is the highest frequency. The resulting dot product decomposes into exact frequency contributions,
\begin{equation}
    \tilde q_t^\top\tilde k_s
    =
    \sum_{r=0}^{F-1}
    \left(q_t^{(r)}\right)^\top
    R\!\left((s-t)\omega_r\right)
    k_s^{(r)},
    \label{eq:rope_logit_decomposition}
\end{equation}
so each group has rotation period $2\pi/\omega_r$ and a distinct positional scale.

\section{Functional Differentiation in RoPE Transformers}
\label{sec:head_functional_differentiation}
\label{sec:method}
\label{sec:rope_functional_differentiation}

\subsection{Behavioral Probes of Attention-Head Tendencies}
\label{subsec:behavioral_detection}

We measure each Query head's retrieval tendency with QRscore, based on query-to-evidence attention~\citep{zhangQueryFocusedRetrievalHeads2025}, and its local-streaming tendency with LDscore, based on the fraction of non-sink attention within a recent window (Appendix~\ref{app:behavioral_probes}).

\begin{figure}[!t]
    \centering
    \includegraphics[width=\linewidth]{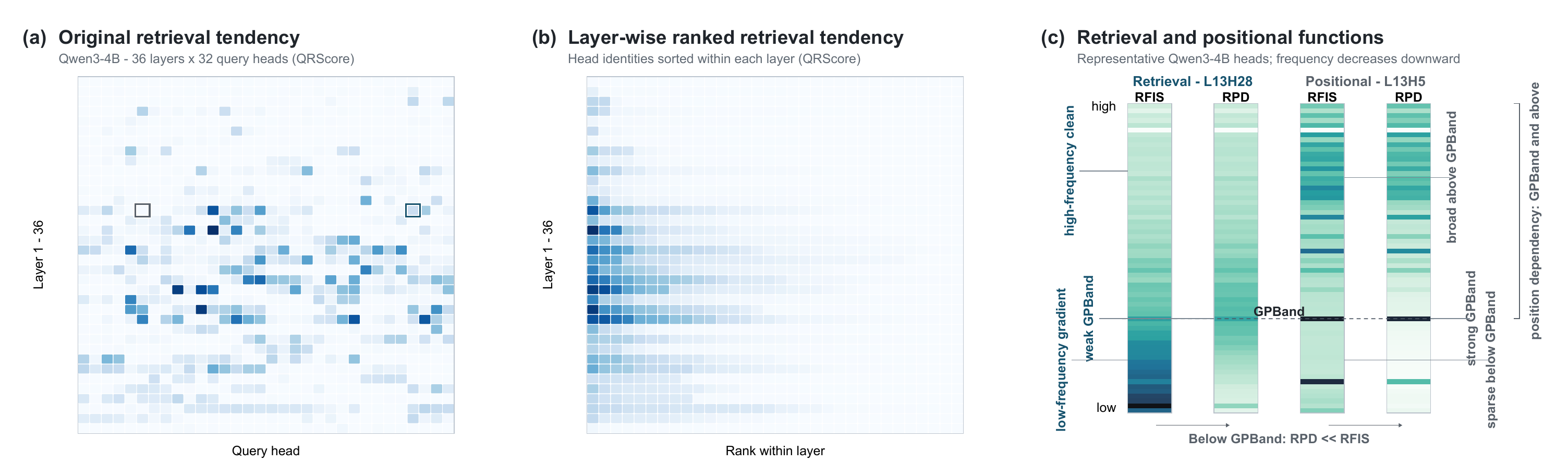}
    \caption{\textbf{Functional differentiation in Qwen3-4B.} \textbf{(a)} Retrieval tendency measured by QRscore. The two boxes mark the heads shown in panel~(c). \textbf{(b)} Layer-specific retrieval distribution after within-layer head sorting. \textbf{(c)} RFIS and RPD for two example heads illustrate retrieval and positional frequency dependence relative to the GPBand.}
    \label{fig:overview}
\end{figure}

Qwen3-4B is the primary model for analysis. Qwen3-1.7B, Qwen3-8B, and Llama3.1-8B-Instruct provide cross-scale and cross-family checks (Appendix~\ref{app:cross_model_consistency}). Its retrieval tendency is sparse, weak in the boundary layers, and concentrated near the middle (Figure~\ref{fig:overview}a--b). Retrieval and local-streaming tendencies provide behavioral anchors and a layer-specific allocation pattern. Their uneven correspondence across layers motivates examining RoPE-frequency dependence directly (Appendix~\ref{app:behavioral_results}).

\subsection{RFIS Reveals Retrieval--Positional Functional Differentiation}
\label{subsec:rfis_analysis}

To characterize the RoPE-frequency dependence underlying these behavioral tendencies, we introduce the \emph{RoPE Frequency Importance Score} (RFIS). RFIS removes one two-dimensional frequency contribution from a head's logits while holding every other contribution fixed, then measures the resulting change in the complete attention distribution. Formally, for sample $x$, let $\mathcal C_x(i)$ be the keys visible to query position $i$. At layer $\ell$, Query head $h$, and frequency $r$, with $g(h)$ its GQA KV head, the exact contribution to the query--key logit is
\begin{equation}
c^{\mathrm{RoPE}}_{\ell,h,i,j,r}
=
s_h\mathbf q_{\ell,h,i,r}^{\top}
R\!\left((j-i)\omega_r\right)
\mathbf k_{\ell,g(h),j,r}.
\label{eq:exact_frequency_contribution}
\end{equation}
Here $s_h$ is the model's attention-logit scale, normally $d_h^{-1/2}$. The complete logits and attention distribution over the visible keys are
\begin{equation}
z_{\ell,h,i,j}
=\sum_{r=0}^{F-1}c^{\mathrm{RoPE}}_{\ell,h,i,j,r},
\qquad
\mathbf p_{\ell,h,i}
=\operatorname{softmax}(\mathbf z_{\ell,h,i}).
\label{eq:complete_frequency_logits}
\end{equation}
Removing frequency $r$ gives
\begin{equation}
\mathbf z_{\ell,h,i}^{(-r)}
=\mathbf z_{\ell,h,i}-\mathbf c^{\mathrm{RoPE}}_{\ell,h,i,r},
\qquad
\mathbf p_{\ell,h,i}^{(-r)}
=\operatorname{softmax}\!\left(\mathbf z_{\ell,h,i}^{(-r)}\right).
\label{eq:rfis_intervention}
\end{equation}
For distributions $\mathbf a,\mathbf b$ and $\mathbf m=(\mathbf a+\mathbf b)/2$, let
\begin{equation}
D_{\mathrm{JS}}(\mathbf a,\mathbf b)
=\frac12D_{\mathrm{KL}}(\mathbf a\Vert\mathbf m)
+\frac12D_{\mathrm{KL}}(\mathbf b\Vert\mathbf m).
\label{eq:js_definition}
\end{equation}
The per-query and aggregate scores are
\begin{equation}
I^{\mathrm{RFIS}}_{\ell,h,i,r}
 =
\frac{D_{\mathrm{JS}}\!\left(
\mathbf p_{\ell,h,i},\mathbf p_{\ell,h,i}^{(-r)}
\right)}{\ln2},
\qquad
\operatorname{RFIS}_{\ell,h,r}
 =\mathbb E_{x\sim\mathcal D}\,
  \mathbb E_{i\sim\mathcal Q(x)}
  \!\left[I^{\mathrm{RFIS}}_{\ell,h,i,r}\right].
\label{eq:rfis}
\end{equation}
The nested expectation weights samples equally. RFIS is a faithful intervention-based metric naturally bounded in $[0,1]$. Larger values indicate stronger dependence on information represented by that frequency. Appendix~\ref{app:theory} gives further theoretical details.

The RFIS panels in Figure~\ref{fig:qwen3_4b_rfis_rpd_qrld} show a salient mid-low-frequency band in both representative layers. Heads with a strong retrieval tendency depend more on frequencies below it, with a general tendency toward greater importance at lower frequencies, and often show little dependence on the band or above. Heads with weaker retrieval tendencies often emphasize the band and higher frequencies. LDscore has no comparably clear relation. Together, these patterns suggest a mechanism-level differentiation between retrieval and positional modeling. However, establishing the positional interpretation of the latter pattern requires isolating rotary modulation, motivating RPD.

\subsection{RPD Verifies the Positional Interpretation of RFIS}
\label{subsec:rpd_analysis}

RFIS identifies important frequency contributions but does not by itself determine whether their effects depend on position. We therefore introduce \emph{RoPE Positional Dependence} (RPD), which replaces the relative rotation of one frequency with the identity while preserving its learned Q/K coordinates:
\begin{equation}
c^{\mathrm{NoPE}}_{\ell,h,i,j,r}
=
s_h\mathbf q_{\ell,h,i,r}^{\top}
\mathbf k_{\ell,g(h),j,r}.
\label{eq:nope_frequency_contribution}
\end{equation}
The resulting single-frequency RoPE-to-NoPE intervention is
\begin{equation}
\mathbf z_{\ell,h,i}^{(r\rightarrow\mathrm{NoPE})}
=
\mathbf z_{\ell,h,i}
-\mathbf c^{\mathrm{RoPE}}_{\ell,h,i,r}
+\mathbf c^{\mathrm{NoPE}}_{\ell,h,i,r},
\qquad
\mathbf p_{\ell,h,i}^{(r\rightarrow\mathrm{NoPE})}
=
\operatorname{softmax}\!\left(
\mathbf z_{\ell,h,i}^{(r\rightarrow\mathrm{NoPE})}
\right).
\label{eq:rpd_intervention}
\end{equation}
We define
\begin{equation}
I^{\mathrm{RPD}}_{\ell,h,i,r}
 =
\frac{D_{\mathrm{JS}}\!\left(
\mathbf p_{\ell,h,i},
\mathbf p_{\ell,h,i}^{(r\rightarrow\mathrm{NoPE})}
\right)}{\ln2},
\qquad
\operatorname{RPD}_{\ell,h,r}
 =\mathbb E_{x\sim\mathcal D}\,
  \mathbb E_{i\sim\mathcal Q(x)}
  \!\left[I^{\mathrm{RPD}}_{\ell,h,i,r}\right].
\label{eq:rpd}
\end{equation}
RPD is likewise a faithful intervention-based metric bounded in $[0,1]$, targeting rotation dependence rather than total channel importance. Measurement settings and further theoretical details are in Appendices~\ref{app:rope_detection_settings} and~\ref{app:theory}.

\begin{figure}[!ht]
    \centering
    \includegraphics[width=\linewidth]{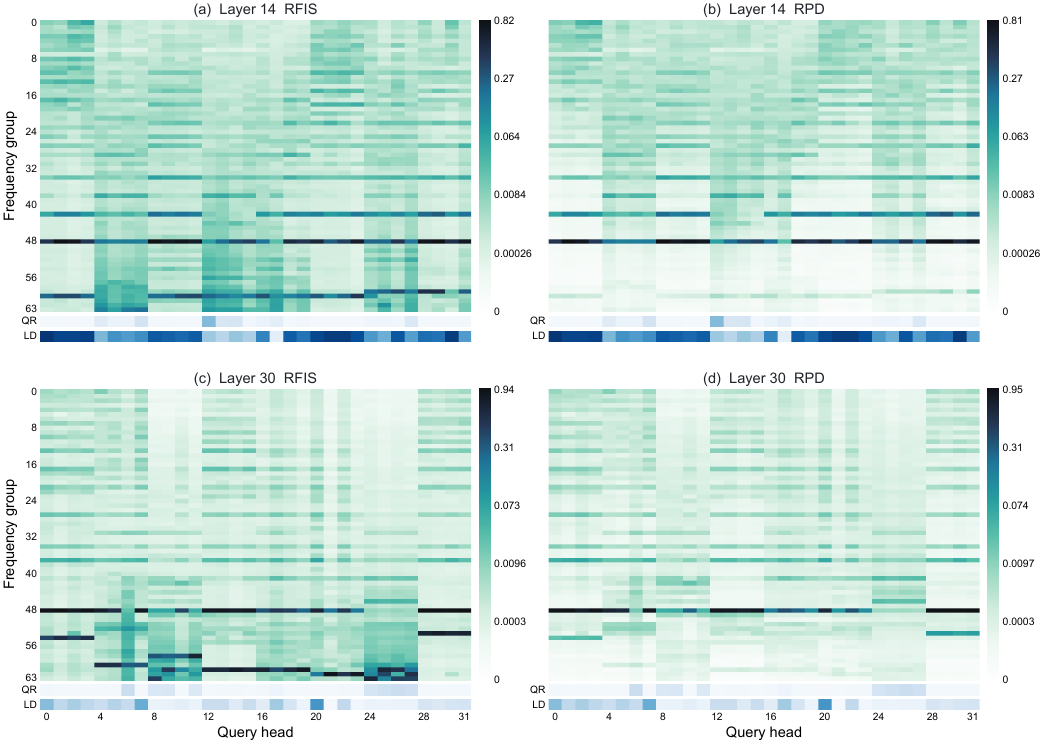}
    \caption{\textbf{RFIS and RPD in Qwen3-4B Layers 14 and 30.} RFIS (left) separates below-band retrieval dependence from band-and-above importance; RPD (right) identifies the latter as positional. QRscore/LDscore strips are aligned below each heatmap; frequencies descend from high to low, and panels use independent zero-to-maximum scales.}
    \label{fig:qwen3_4b_rfis_rpd_qrld}
\end{figure}

RPD follows RFIS at the salient band and above, confirming that this pattern depends on rotary positional modulation (Figures~\ref{fig:overview}c and~\ref{fig:qwen3_4b_rfis_rpd_qrld}, right). Below the band, RPD is generally much weaker, consistent with slow rotations close to the identity over the observed offsets. Heads with a strong retrieval tendency can depend substantially on these frequencies as measured by RFIS. RFIS and RPD thus reveal a complete functional differentiation between retrieval and positional modeling at the level of RoPE-frequency dependence.

The same trends appear in our smaller pretrained Transformers (Appendix~\ref{app:controlled_boundary}, \hyperref[par:pretrained_functional_differentiation]{\emph{Functional differentiation in our pretrained Transformers}}). Additional Qwen3, Llama3.1, and Gemma analyses appear in Appendix~\ref{app:rope_results}.

\paragraph{Principle 1. Functional organization.}
This implicitly learned differentiation does not amount to strict head-level separation: many individual heads exhibit characteristics of both patterns. We therefore hypothesize that explicitly separating these functions may improve autoregressive language modeling. Together with the coexistence of the two functional tendencies within layers and the variation in retrieval tendency across depth, this motivates the first principle: \emph{Retrieval and positional functions should be assigned separately and mixed within layers with layer-specific allocation.}

\subsection{Global Positional Band}
\label{subsec:controlled_boundary}

Because the salient band separates positional dependence from approximately position-invariant retrieval dependence, we hypothesize that its positional scale is tied to the context length over which global positional structure is learned. Production LLMs with context-extension training can exhibit multiple depth-varying bands (Appendix~\ref{app:qwen_layermean}). We therefore train two $380$M-parameter QKNorm Transformers at sequence length $2048$, with RoPE bases $\theta=10^4$ and $10^6$. A third model uses a training length of $4096$ at $\theta=10^4$.

\begin{figure}[!htbp]
    \centering
    \includegraphics[width=\linewidth]{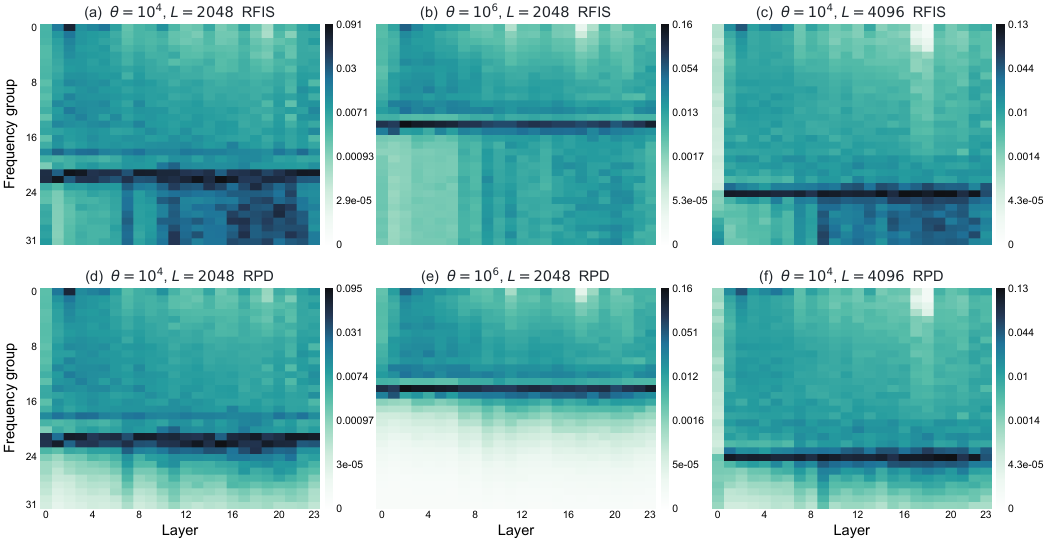}
    \caption{\textbf{Frequency dependence under controlled training lengths and RoPE bases.} The top row shows RFIS and the bottom row RPD, each averaged over Query heads within each layer. Columns show $(\theta,L_{\mathrm{train}})=(10^4,2048)$, $(10^6,2048)$, and $(10^4,4096)$. The diagnostic length matches the training length in each column. Each panel uses its own zero-to-maximum color scale.}
    \label{fig:controlled_boundary}
\end{figure}

At training length $2048$, the band lies at groups $21$--$22$ for $\theta=10^4$ (periods $2650$ and $3533$ tokens) and at groups $14$--$15$ for $\theta=10^6$ (periods $2650$ and $4080$ tokens). At $\theta=10^4$, the model trained at length $4096$ peaks near group $24$, with a period of $6283$ tokens (Figure~\ref{fig:controlled_boundary}). We refer to this training-length-related boundary between retrieval and positional dependence as the \emph{Global Positional Band} (GPBand). These locations agree with the criterion of \citet{okaFREQUENCYBANDSROPE2026}, which predicts the frequency that maximizes the variance of a fixed-norm linear combination of the two RoPE coordinates over the training length. Appendix~\ref{app:controlled_boundary} gives the derivation and further comparisons.

\paragraph{Masking the band after training.}
We replace only the GPBand rotations (groups $21$--$22$) with the identity in the trained full-RoPE model, removing positional modulation while preserving all Q/K content dimensions and learned weights. Mean cloze retrieval within the $2$K budget falls from $27.72\%$ to $21.57\%$, while short-context commonsense accuracy remains nearly unchanged ($42.66\%$ versus $42.63\%$). Among the only three tasks containing original prompts longer than $2048$ tokens and truncated to fit the $2$K evaluation budget, FDA drops from $17.60\%$ to $2.18\%$, SWDE from $29.34\%$ to $17.64\%$, and NQ from $14.67\%$ to $8.77\%$. Other adjacent-pair masks sampled at evenly spaced group indices show no comparable collapse on these tasks (Appendix~\ref{app:gpband_inference_ablation}). Together with the training-length-dependent band location, these losses support the conclusion that models with global positional modeling fit the training sequence length.

\begin{table}[!t]
    \centering
    \footnotesize
    \setlength{\tabcolsep}{3.1pt}
    \caption{\textbf{Restricting RoPE frequencies during pretraining.} All three models use QKNorm and a $2$K training length. The retained group range is inclusive, and \emph{full} retains all $32$ groups. Commonsense and retrieval are task means in percent, with a $2$K budget for real-world retrieval. RULER NIAH entries are accuracy in percent. Appendix~\ref{app:gpband_extrapolation} shows the same models under the dense length--depth protocol.}
    \label{tab:gpband_pretraining}
    \begin{tabular*}{\linewidth}{@{\extracolsep{\fill}}lrrrrrrrrrrr@{}}
        \toprule
        & & & \multicolumn{3}{c}{NIAH-Single-1} & \multicolumn{3}{c}{NIAH-Single-2} & \multicolumn{3}{c}{NIAH-Single-3} \\
        \cmidrule(lr){4-6}\cmidrule(lr){7-9}\cmidrule(l){10-12}
        RoPE & Commonsense & Retrieval & 1K & 2K & 4K & 1K & 2K & 4K & 1K & 2K & 4K \\
        \midrule
        full & 42.66 & 27.72 & 100.0 & 77.8 & 0.2 & 15.2 & 47.4 & 11.0 & 20.0 & 51.4 & 2.8 \\
        $0$--$22$ & 42.49 & 30.78 & 100.0 & 99.8 & 0.0 & 20.8 & 89.8 & 11.8 & 14.0 & 45.2 & 0.0 \\
        $0$--$20$ & 42.63 & 30.24 & 99.8 & 99.8 & 98.2 & 100.0 & 100.0 & 83.2 & 96.0 & 95.4 & 26.6 \\
        \bottomrule
    \end{tabular*}
\end{table}

\paragraph{Withholding the band during pretraining.}
This intervention tests how the availability of GPBand rotations during learning affects extrapolation. We compare full RoPE with two pretrained variants that both disable slower rotations: $0$--$22$ retains GPBand rotations, whereas $0$--$20$ withholds them. Withholding the band markedly improves single-needle retrieval beyond the training length, with comparable commonsense reasoning and real-world retrieval performance relative to $0$--$22$ (Table~\ref{tab:gpband_pretraining}). The improvement thus occurs without a broad performance collapse within the training-length budget, supporting the interpretation that learned training-scale positional dependence can impair extrapolation. The dense length--depth evaluation nevertheless reveals nonmonotonic failures, including at $3$K (Appendix~\ref{app:gpband_extrapolation}). Higher-frequency rotations may jointly support residual global positional fitting: withholding GPBand leaves positional modeling with global access and does not ensure stable extrapolation.

\FloatBarrier

\paragraph{Principle 2. Positional dependence.}
The above analysis indicates that RoPE Transformers fit the training scale. This global positional fitting may limit generalization to longer sequences. It motivates the second principle. \emph{Positional modeling should operate only locally, while global access should be implemented through position-independent retrieval.}

\section{Testing the Two Principles}
\label{sec:hwh_evaluation}
\label{sec:experiments}
\label{subsec:hwh_design}

We test the principles from Section~\ref{sec:rope_functional_differentiation} using matched models trained from scratch on FineWeb-Edu~\citep{penedoFineWebDatasetsDecanting2024} with a $2$K context window. We use $15$B training tokens at $380$M parameters and $100$B at $1.4$B. The three reference architectures are Transformer with RoPE FA, GDN with pure LA, and the layer-wise hybrid \emph{Inter}, which alternates three GDN layers with one NoPE FA layer. The Head-wise Hybrid Architecture (HwH) enables flexible assignment of mechanisms to individual heads, concatenating their outputs before the output projection. Appendix~\ref{app:experimental_details} gives training and evaluation details.

\paragraph{Testing the principles at $380$M.}
Table~\ref{tab:principle_ablation} examines five aspects of the principles. Principle~1 calls for (1) separate components for retrieval and positional modeling, (2) mixing within layers, and (3) layer-specific allocation. Principle~2 calls for (4) local positional modeling and (5) position-independent global retrieval. We implement (5) with NoPE FA. Although causal contextualization can still convey positional information~\citep{haviv-etal-2022-transformer}, we adopt NoPE FA as the most direct implementation of position-independent global retrieval within the autoregressive framework, as its attention scores introduce no explicit positional modulation. GDN approximates (4) through a decayed recurrent state that can nevertheless carry information from distant tokens. Sink-free RoPE sliding-window attention (SWA) instead enforces a hard $128$-token window on direct positional interactions, avoiding the implicit global positional modeling that can arise from accumulated decay in LA's recurrent state.

\begin{table}[!hb]
\centering
\caption{Main architectural ablations for testing the two design principles at $380$M. Two checks in (3) indicate reduced retrieval allocation in shallow and deep layers in addition to its removal from the first and last layers. In (4), one and two checks denote decay-based locality and a hard local window, respectively. The NIAH-Single-2 column averages accuracy across the $15$ lengths and $11$ insertion depths in Figure~\ref{fig:niah_all_380m}.}
\label{tab:principle_ablation}
\small
\setlength{\tabcolsep}{3.3pt}
\begin{tabular*}{\linewidth}{@{\extracolsep{\fill}}lccccc rrr@{}}
\toprule
& \multicolumn{3}{c}{Principle 1} & \multicolumn{2}{c}{Principle 2} & \multicolumn{3}{c}{Mean scores (\%, $\uparrow$)} \\
\cmidrule(lr){2-4}\cmidrule(lr){5-6}\cmidrule(l){7-9}
Model & (1) & (2) & (3) & (4) & (5) & Commonsense & Retrieval & NIAH-Single-2 \\
\midrule
Transformer & $\times$ & $\times$ & $\times$ & $\times$ & $\times$ & 42.30 & 24.95 & 13.82 \\
GDN         & $\times$ & $\times$ & $\times$ & $\checkmark$ & $\times$ & 42.18 & 20.02 & 42.36 \\
Inter       & $\checkmark$ & $\times$ & $\times$ & $\checkmark$ & $\checkmark$ & 42.07 & 28.97 & 49.52 \\
HwH-uni     & $\checkmark$ & $\checkmark$ & $\times$ & $\checkmark$ & $\checkmark$ & 42.93 & 27.10 & 49.39 \\
HwH-rmfl    & $\checkmark$ & $\checkmark$ & $\checkmark$ & $\checkmark$ & $\checkmark$ & 42.88 & 26.99 & 60.06 \\
HwH-std     & $\checkmark$ & $\checkmark$ & $\checkmark\checkmark$ & $\checkmark$ & $\checkmark$ & 42.78 & 28.98 & 68.42 \\
HwH-std-r   & $\checkmark$ & $\checkmark$ & $\checkmark\checkmark$ & $\times$ & $\times$ & 42.64 & 23.67 & 22.42 \\
HwH-swa     & $\checkmark$ & $\checkmark$ & $\checkmark\checkmark$ & $\checkmark\checkmark$ & $\checkmark$ & 42.13 & 29.32 & 89.88 \\
\bottomrule
\end{tabular*}
\end{table}

\begin{figure}[!hb]
\centering
\includegraphics[width=\linewidth]{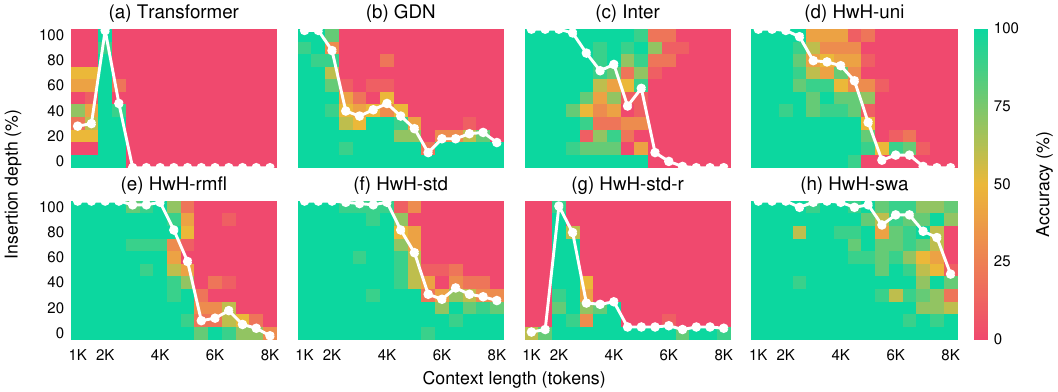}
\caption{\textbf{NIAH-Single-2 heatmaps at $380$M.} Context length increases from $1$K to $8$K in $512$-token steps. Depth runs from $0\%$ at the sequence end (bottom) to $100\%$ at the beginning (top), with $10$ examples per cell. White curves mark depth-averaged accuracy on the same $0$--$100\%$ scale. \hyperref[app:dense_niah_robustness]{Appendix~\ref*{app:experimental_details}} discusses the unusually low $1$K scores of Transformer and HwH-std-r.}
\label{fig:niah_all_380m}
\end{figure}

\paragraph{Principle 1. Explicit functional separation and layer-specific within-layer mixing.}
\label{subsec:allocation_ablation}

Inter improves mean real-world retrieval and improves length extrapolation over Transformer while retaining comparable commonsense accuracy (Table~\ref{tab:principle_ablation}). By separating the functions across layers and combining GDN with NoPE FA, it partially implements both principles. Its gains provide initial support for explicit functional separation coupled with local positional modeling and position-independent global retrieval.

HwH-uni redistributes the same components and overall FA budget within layers. Compared with Inter, it improves commonsense accuracy but yields no extrapolation gain and lower real-world retrieval. Thus, uniform within-layer mixing alone does not account for the benefits of the subsequent designs.

Layer-specific allocation provides the additional gains. Removing retrieval components from the first and last layers in HwH-rmfl raises depth-averaged NIAH-Single-2 accuracy at $4$K (Figure~\ref{fig:niah_all_380m}d--f). HwH-std further follows the observed depth-dependent retrieval profile by reducing retrieval allocation in shallow and deep layers outside the middle half. It matches HwH-rmfl at $4$K, improves extrapolation beyond $5$K, and raises mean real-world retrieval despite a reduced total FA budget. Its retrieval mean essentially matches Inter while substantially exceeding Transformer. The GDN-based HwH variants also exceed Transformer in mean commonsense accuracy at this scale. These results support Principle~1 as a combination of explicit functional separation and layer-specific within-layer mixing. Appendix~\ref{app:learn_allocation} discusses learning that allocation during pretraining.

\paragraph{Principle 2. Local positional modeling and position-independent global retrieval.}
\label{subsec:component_ablation}

Replacing NoPE FA with RoPE FA in HwH-std-r sharply degrades retrieval and extrapolation while leaving commonsense accuracy nearly unchanged. This intervention preserves component allocation but reintroduces global positional modeling and position-dependent retrieval, supporting the importance of keeping explicit positional modulation out of the global mechanism.

HwH-swa replaces GDN with SWA, removing long-range recurrence from the positional component. Extrapolation improves substantially and real-world retrieval improves modestly. The drop in commonsense accuracy shows that mechanism choice also matters for general modeling. Appendix~\ref{app:mixing_position} further explains why enforcing local positional dependence more strictly can improve length extrapolation.

\begin{table}[!hb]
\centering
\caption{Language modeling perplexity ($\downarrow$) and zero-shot commonsense reasoning accuracy (\%, $\uparrow$) at $1.4$B parameters. Best and second-best results are bold and underlined, respectively.}
\label{tab:main_lm_reasoning}
\scriptsize
\setlength{\tabcolsep}{3.4pt}
\resizebox{\linewidth}{!}{%
\begin{tabular}{lrr|rrrrrrrr}
\toprule
Model & Wiki $\downarrow$ & LAMB. $\downarrow$ & LAMB. & HellaS. & PIQA & ARC-E & ARC-C & WinoG. & OBQA & Avg. \\
\midrule
Transformer & 17.36 & 13.31 & \underline{46.30} & 55.30 & \textbf{72.42} & 69.87 & \underline{38.31} & 56.91 & \textbf{29.20} & 52.61 \\
GDN         & 17.32 & 13.42 & 44.19 & 55.54 & 72.20 & \underline{71.30} & 36.69 & \textbf{58.96} & 27.60 & 52.35 \\
Inter       & \textbf{16.82} & \textbf{11.97} & \textbf{47.68} & \textbf{55.99} & \underline{72.36} & \textbf{71.55} & \textbf{38.48} & \underline{58.48} & 26.80 & \textbf{53.05} \\
HwH-std     & \underline{16.85} & \underline{12.11} & \textbf{47.68} & \underline{55.96} & 72.09 & 70.12 & 36.60 & 58.25 & \underline{28.20} & \underline{52.70} \\
HwH-swa     & 17.54 & 14.74 & 44.03 & 55.12 & 71.76 & 69.78 & 37.63 & 57.46 & 26.00 & 51.68 \\
\bottomrule
\end{tabular}%
}
\end{table}

\begin{table}[!htb]
\centering
\caption{Real-world retrieval and RULER NIAH at the $1.4$B scale. }
\label{tab:main_retrieval}
\scriptsize
\setlength{\tabcolsep}{2.4pt}
\resizebox{\linewidth}{!}{%
\begin{tabular}{lrrrrrrrrrrrrrrrr}
\toprule
& \multicolumn{7}{c}{Real-world retrieval} & \multicolumn{3}{c}{NIAH-Single-1} & \multicolumn{3}{c}{NIAH-Single-2} & \multicolumn{3}{c}{NIAH-Single-3} \\
\cmidrule(lr){2-8}\cmidrule(lr){9-11}\cmidrule(lr){12-14}\cmidrule(lr){15-17}
Model & FDA & SWDE & SQuAD & NQ & TriviaQA & DROP & Avg. & 1K & 2K & 4K & 1K & 2K & 4K & 1K & 2K & 4K \\
\midrule
Transformer & 46.19 & \underline{45.27} & 40.68 & 25.12 & \textbf{61.26} & 22.47 & 40.17 & \textbf{100.0} & \textbf{100.0} & 0.0 & \textbf{100.0} & \underline{99.8} & 0.0 & 70.8 & \underline{67.6} & 0.0 \\
GDN         & 16.33 & 22.95 & 35.22 & 21.79 & 58.23 & 21.42 & 29.33 & \textbf{100.0} & \textbf{100.0} & 99.4 & \underline{99.8} & 85.0 & 35.2 & \underline{87.0} & 52.6 & 23.0 \\
Inter       & \underline{64.43} & 40.95 & 42.73 & \textbf{26.64} & 59.72 & \underline{24.01} & 43.08 & \textbf{100.0} & \textbf{100.0} & \underline{99.8} & \textbf{100.0} & \textbf{100.0} & 79.6 & 39.8 & 67.4 & 35.8 \\
HwH-std     & \textbf{66.15} & \textbf{47.25} & \underline{45.84} & \underline{25.28} & 60.13 & 23.19 & \textbf{44.64} & \textbf{100.0} & \textbf{100.0} & \textbf{100.0} & \underline{99.8} & \textbf{100.0} & \textbf{99.6} & 69.2 & 58.4 & \underline{46.2} \\
HwH-swa     & 55.63 & 44.28 & \textbf{48.86} & \textbf{26.64} & \underline{60.19} & \textbf{24.58} & \underline{43.36} & \textbf{100.0} & \textbf{100.0} & 90.8 & \textbf{100.0} & \textbf{100.0} & \underline{98.8} & \textbf{99.2} & \textbf{96.2} & \textbf{89.6} \\
\bottomrule
\end{tabular}%
}
\end{table}

\begin{figure}[!ht]
\centering
\includegraphics[width=\linewidth]{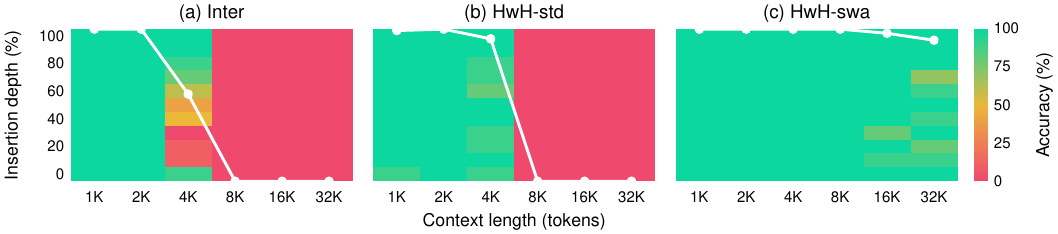}
\caption{\textbf{NIAH-Single-2 heatmaps at $1.4$B with inference-time query scaling.} All three models use SWAN-style logarithmic query scaling~\citep{puvvadaSWANEfficientScalable2025a} ($\alpha=300$) in NoPE FA. Depth and white curves follow Figure~\ref{fig:niah_all_380m}.}
\label{fig:niah_extrapolation_1p4b}
\end{figure}

\paragraph{Consistency across model scales.}
At $1.4$B, Inter and HwH-std retain competitive language modeling and commonsense reasoning while mean real-world retrieval rises from $40.17\%$ (Transformer) to $43.08\%$ (Inter) and $44.64\%$ (HwH-std; Tables~\ref{tab:main_lm_reasoning}--\ref{tab:main_retrieval}). Appendix~\ref{app:retrieval_space} explains how explicit functional separation frees the full retrieval-head width for content matching. With identical query scaling, HwH-swa sustains extrapolation through $32$K while Inter and HwH-std fail from $8$K (Figure~\ref{fig:niah_extrapolation_1p4b}; Appendix~\ref{app:experimental_details}). Its weaker short-context scores highlight mechanism choice.

The additional $380$M Inter-swa control improves on Inter through $4.5$K, then falls to $33.6\%$ at $5$K and $10.9\%$ at $6$K. HwH-swa reaches $90.0\%$ at $6$K with the same $128$-token window (Appendix~\ref{app:inter_swa_ablation}). This supports layer-specific within-layer mixing for sustained extrapolation, consistent with Principle~1.

Inter leads in commonsense accuracy at $1.4$B, unlike at $380$M. HwH-std exceeds Transformer at both scales. Inter and HwH-std both extrapolate better than Transformer (Appendix~\ref{app:mixing_position}). Their real-world retrieval scores are tied at $380$M, and HwH-std leads at $1.4$B. Thus, implementing the principles improves retrieval and extrapolation while preserving competitive general modeling.

At $4$K, both scales' heatmaps show middle-depth failures in Inter and more uniform retrieval in HwH-std; HwH-swa extends this robust region. This suggests a possible connection to mitigating lost-in-the-middle behavior~\citep{liuLostMiddleHow2024}, to investigate beyond synthetic single-needle retrieval.

Pure GDN trails Transformer in mean commonsense and real-world retrieval at both scales, despite some perplexity and extrapolation advantages. Its stronger combined performance alongside NoPE global retrieval supports complementary functions and their organization, rather than a uniformly superior standalone mechanism.

\FloatBarrier

\section{Related Works}
\label{sec:related_works}

\paragraph{Mechanistic Analysis of Attention Heads and RoPE.}

\citet{wuRetrievalHeadMechanistically2024} identify retrieval heads and causally link them to long-context recall; \citet{xiaoDuoAttentionEfficientLongContext2024} detect them with trainable gates and assign other heads to streaming; \citet{zhangQueryFocusedRetrievalHeads2025} extend detection to query-to-evidence behavior; and \citet{tanHydraHeadHeadLevelFunctional2026} assess head necessity through activation and path patching. \citet{hongTokenDistanceModeling2024} connect non-uniform RoPE frequency use to token distance. \citet{barberoWeGoWhat2024} use Q/K norms to link high frequencies to position and low frequencies to semantic matching. \citet{urrutiaDecouplingPositionalSymbolic2025} derive per-frequency positional and symbolic scores through input permutation. These accounts leave the role of the middle frequencies open. \citet{chiang-yogatama-2025-rotary} find that fast-rotating coordinates can be underused for distant retrieval. Using Q/K norms, \citet{okaFREQUENCYBANDSROPE2026} observe the salient band and relate its location theoretically to the RoPE base and training length.

\paragraph{Hybrid Architectures.}
Hybrid architectures often target a favorable efficiency--quality balance by combining attention with compact recurrent memory~\citep{lieberJambaHybridTransformerMamba2024,wangSystematicAnalysisHybrid2025}. Some designs also explicitly exploit complementary memory roles, including context summarization and fine-grained retrieval~\citep{dongHymbaHybridheadArchitecture2024,zancatoBMOJOHybridState2024,nunezExpansionSpanCombining2024}.

Existing designs span several allocation granularities. Layer-wise hybrids interleave FA with LA or state-space layers~\citep{lieberJambaHybridTransformerMamba2024,minimaxMinimax01ScalingFoundation2025,waleffeEmpiricalStudyMamba2024,teamKimiLinearExpressive2025,QwenQwen3627BHugging2026}. \citet{wangSystematicAnalysisHybrid2025} systematically study these allocations, while \citet{chenHybridLinearAttention2026} refine hybrid architectures and positional encoding when converting Qwen3 models. Within-layer designs combine local and global computations~\citep{huaTransformerQualityLinear2022} or parallel attention and state-space components~\citep{dongHymbaHybridheadArchitecture2024,zuoFalconH1FamilyHybridHead2025}. \citet{tanHydraHeadHeadLevelFunctional2026} preserve FA at retrieval-critical heads during checkpoint conversion, whereas \citet{liTransMambaSequenceLevelHybrid2026} switch mechanisms across token positions and layers.

\citet{munkhdalaiLeaveNoContext2024} and \citet{zhangLoLCATsLowrankLinearizing2025} use LA to retain distant history alongside local softmax attention. \citet{yang2025rnope} identify a division of labor across NoPE and RoPE layers and improve extrapolation by restricting only the RoPE layers to local windows. \citet{teamKimiLinearExpressive2025} pairs global NoPE with position-sensitive LA whose recurrent state can retain distant information. Our RFIS/RPD analysis traces retrieval and positional modeling to distinct frequency dependencies around GPBand in RoPE Transformers. From this implicit differentiation we derive principles for where to assign each function and which function should have global access. These principles account for benefits of existing designs and motivate developing LA for expressive local positional modeling alongside separate global retrieval.

\section{Conclusion}
\label{sec:conclusion}

RFIS and RPD establish a complete mechanism-level taxonomy of retrieval and positional modeling in RoPE Transformers, with the training-length-related GPBand separating their frequency dependencies. Individual heads can mix both functions. Post-training and pretraining GPBand interventions further show that models with global positional modeling fit the training span, and that this learned dependence can impair length extrapolation. These findings motivate two principles for autoregressive language modeling: \textbf{(1) Functional Organization:} retrieval and positional functions should be assigned separately and mixed within layers with layer-specific allocation; and \textbf{(2) Positional Dependence:} positional modeling should operate only locally, while global access should rely on position-independent retrieval. From-scratch pretraining at $380$M and $1.4$B, together with targeted ablations, supports the principles.

The four implications progress from practical applications to directions for future research:
\begin{enumerate}
\item \textbf{Practical tools and architecture.} Apply RFIS/RPD to Transformer mechanism analysis. Derive practical rules for functional allocation or develop strategies that let models learn their own allocations during pretraining (Appendix~\ref{app:learn_allocation}).
\item \textbf{Understanding existing gains.} Explain hybrid architectures' advantages over Transformer. Reduced middle-depth failures may also offer clues for investigating lost-in-the-middle.
\item \textbf{Promising research directions.} Pursue better extrapolation by abandoning global positional modeling. Develop brand-new Linear Attention and related recurrent mechanisms for expressive local positional modeling, for example, by developing parallelizable finite-window nonlinear local modeling mechanisms, alongside separate global retrieval.
\item \textbf{Functional hybridization.} Pursue architectures for complementary functions, beyond an efficiency--quality tradeoff. Better position-independent global retrieval and local positional modeling and their hybridization scheme may support designs beyond today's stacked multi-head attention template.
\end{enumerate}

\section*{Reproducibility Statement}
We provide the implementation and experiment configurations in the supplementary material, with a corresponding anonymous repository available at \url{https://anonymous.4open.science/r/Two-Principles-for-Autoregressive-Language-Modeling}. Future updates to the repository may include model checkpoints (e.g., Hugging Face paths), detailed QR/LD/RFIS/RPD results and visualizations, and synchronized code revisions. Detailed settings for the behavioral probes, RFIS and RPD analyses, controlled GPBand experiments, model pretraining, evaluation protocols, and HwH variants are documented in the corresponding sections of the Appendix. All architecture comparisons at each model scale use matched training data, optimization schedules, and evaluation protocols to facilitate controlled reproduction of the reported results.

\section*{AI Use Statement}
Generative AI tools were used to assist with code implementation, literature retrieval, drafting and revising text, \LaTeX{} formatting, and figure preparation. The authors reviewed all AI-assisted outputs, tested the AI-assisted code, and take full responsibility for the final manuscript.

\bibliography{main}
\bibliographystyle{iclr2027_conference}

\appendix

\section{Detailed Behavioral Analysis}
\label{app:head_analysis}

\subsection{Behavioral Probe Definitions and Measurement Settings}
\label{app:behavioral_probes}

\paragraph{QRscore.}
We follow the query-focused retrieval-head detection method of \citet{zhangQueryFocusedRetrievalHeads2025}.

Consider a prompt $\{\mathcal D,q\}$, where $\mathcal D=\{d_1,d_2,\ldots,d_N\}$ is a sequence of candidate passages followed by query $q$, and let $\mathcal D^*(q)\subseteq\mathcal D$ be the gold-passage set. For layer $\ell$ and Query head $h$, let $A_{\ell,h}\in\mathbb{R}^{T\times T}$ denote the post-softmax causal attention matrix, and let $\mathcal Q(q)$ and $\mathcal D_i$ denote the token positions of $q$ and $d_i$, respectively. Following the original definition, the query-focused retrieval score toward passage $d_i$ is
\begin{equation}
\operatorname{QRscore}_{\ell,h}(q,d_i)
=
\frac{1}{|\mathcal Q(q)|}
\sum_{u\in\mathcal Q(q)}
\sum_{v\in\mathcal{D}_i}
A_{\ell,h}(u,v).
\label{eq:qr_passage_score}
\end{equation}
Aggregating over the gold passages gives the score for query $q$:
\begin{equation}
\operatorname{QRscore}_{\ell,h}(q)
=
\sum_{d_i\in\mathcal D^*(q)}
\operatorname{QRscore}_{\ell,h}(q,d_i).
\label{eq:qr_sample_score}
\end{equation}
For an evaluation dataset $\mathcal T=\{(q,\mathcal D,\mathcal D^*(q))\}$, the head-level score is
\begin{equation}
\operatorname{QRscore}_{\ell,h,\mathcal T}
=
\frac{1}{|\mathcal T|}
\sum_{(q,\mathcal D,\mathcal D^*(q))\in\mathcal T}
\operatorname{QRscore}_{\ell,h}(q).
\label{eq:qr_dataset_score}
\end{equation}
We abbreviate $\operatorname{QRscore}_{\ell,h,\mathcal T}$ as $\operatorname{QRscore}_{\ell,h}$ when the evaluation dataset is fixed.

\paragraph{LDscore.}
We introduce LDscore to measure local-streaming tendency: the extent to which a head concentrates its non-sink attention within a recent context window. For a long-context sample $x$, let $\mathcal U(x)$ denote the evaluated query-token positions. We use zero-based token indices and exclude position $0$, which often serves as an attention sink. For $u\in\mathcal U(x)$, recent-window size $R$, layer $\ell$, and Query head $h$, the token-level score is
\begin{equation}
\operatorname{LDscore}_{\ell,h}(u;R)
=
\frac{
\sum_{v=\max(1,u-R+1)}^{u} A_{\ell,h}(u,v)
}{
\sum_{v=1}^{u} A_{\ell,h}(u,v)
}.
\label{eq:local_dependence_token}
\end{equation}
Averaging over the evaluated query positions gives the sample-level score:
\begin{equation}
\operatorname{LDscore}_{\ell,h}(x;R)
=
\frac{1}{|\mathcal U(x)|}
\sum_{u\in\mathcal U(x)}
\operatorname{LDscore}_{\ell,h}(u;R),
\label{eq:local_dependence_sample}
\end{equation}
For an evaluation dataset $\mathcal X$, the head-level score is
\begin{equation}
\operatorname{LDscore}_{\ell,h,\mathcal X}(R)
=
\frac{1}{|\mathcal X|}
\sum_{x\in\mathcal X}
\operatorname{LDscore}_{\ell,h}(x;R).
\label{eq:local_dependence_dataset}
\end{equation}
We abbreviate $\operatorname{LDscore}_{\ell,h,\mathcal X}(R)$ as $\operatorname{LDscore}_{\ell,h}$ when the evaluation dataset and window size are fixed.

\paragraph{Measurement and visualization settings.}
All behavioral measurements are computed at Query-head granularity. Inputs are truncated to $32{,}768$ tokens for the Qwen3 series and Llama3.1-8B-Instruct, and to $8{,}192$ tokens for Gemma-7B. The $32{,}768$-token setting matches Qwen3's long-context pretraining length~\citep{yangQwen3TechnicalReport2025}. These QRscore measurements use $64$ LongMemEval \texttt{single-session-user} examples with nonempty gold evidence. For the $32$K QRscore inputs, truncation preserves the relative position of the gold evidence; Gemma inputs preserve passage order while shortening passages to fit $8$K. For LDscore, we use $R=128$, sample $100$ long-context sequences from the \texttt{book-65536} subset of ProLong~\citep{gaoHowTrainLongContext2025}, tokenize them with the corresponding model tokenizer, truncate each sequence to the model-specific length above, and evaluate the final $128$ retained query positions. For visualization, each model--metric matrix is min--max normalized independently; the analyses use the original scores.

\subsection{Qwen3-4B Behavioral Patterns}
\label{app:behavioral_results}

\begin{figure}[htbp]
    \centering
    \includegraphics[width=0.96\linewidth]{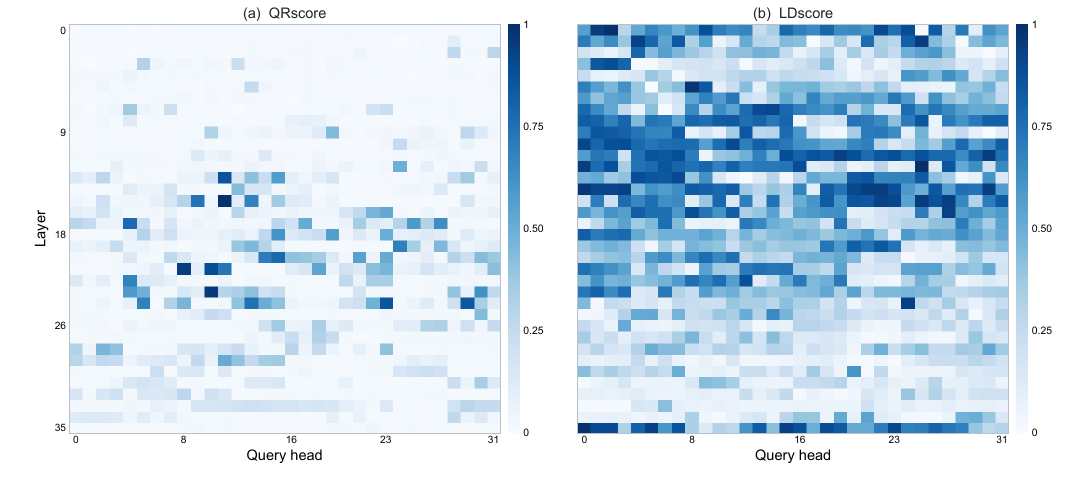}
    \caption{\textbf{Qwen3-4B behavioral tendencies.} \textbf{(a)} Retrieval tendency measured by QRscore. \textbf{(b)} Local-streaming tendency measured by LDscore. Rows denote layers and columns Query heads; each metric is independently normalized for visualization.}
    \label{fig:behavioral_detection}
\end{figure}

In Figure~\ref{fig:behavioral_detection}, the first and last layers show strong local-streaming tendency and little retrieval tendency. The two tendencies are approximately complementary in parts of the middle layers. Low QRscores in layers $1$--$6$ accompany heterogeneous LDscores, and both tendencies are generally weaker in layers $25$--$34$. This uneven correspondence motivates measuring the frequency dependence of the attention distributions directly.

\subsection{Cross-Scale and Cross-Family Behavioral Consistency}
\label{app:cross_model_consistency}

\begin{figure}[!ht]
    \centering
    \includegraphics[width=\linewidth]{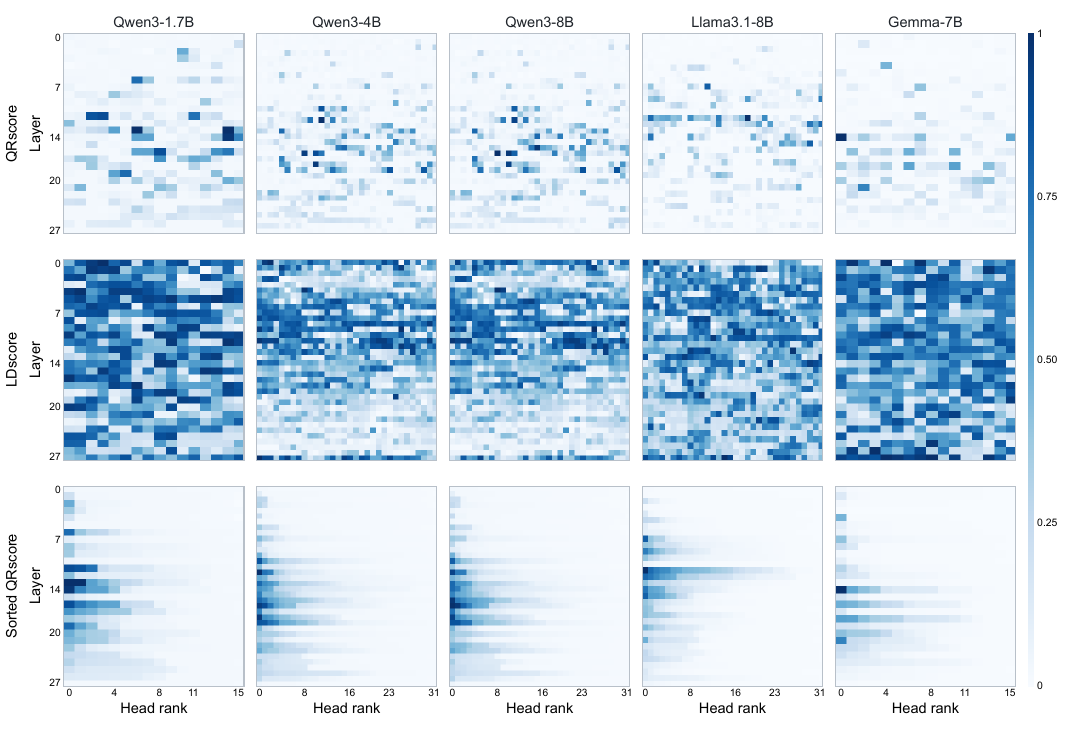}
    \caption{\textbf{Behavioral tendencies across model scales and families.} Columns show Qwen3-1.7B, Qwen3-4B, Qwen3-8B, Llama3.1-8B-Instruct, and Gemma-7B. Rows show retrieval tendency (QRscore), local-streaming tendency (LDscore), and retrieval tendency sorted in descending order within each layer. The first two rows preserve the original Query-head order; each model--metric matrix is independently normalized for visualization.}
    \label{fig:behavioral_cross_model}
\end{figure}

Figure~\ref{fig:behavioral_cross_model} shows consistent retrieval and local-streaming trends across the three Qwen3 scales~\citep{yangQwen3TechnicalReport2025}, with similar layer-specific organization in Llama3.1-8B-Instruct~\citep{grattafioriLlama3Herd2024} and Gemma-7B~\citep{teamGemmaOpenModels2024}.

\FloatBarrier

\section{Additional RFIS and RPD Analysis}
\label{app:rope_results}

\subsection{Measurement Settings}
\label{app:rope_detection_settings}

For the pretrained-model RFIS and RPD measurements, we sample $100$ long-context sequences from the \texttt{book-65536} subset of ProLong~\citep{gaoHowTrainLongContext2025} and tokenize them with the corresponding model tokenizer. Sequences are truncated to $32{,}768$ tokens for the Qwen3 series and Llama3.1-8B-Instruct, and to $8{,}192$ tokens for Gemma-7B. For the controlled $380$M Transformers, diagnostic sequences are truncated to the corresponding training length of $2{,}048$ or $4{,}096$ tokens. In all cases, both metrics are evaluated over the final $128$ retained query positions.

We cache the pre-RoPE Q/K vectors, reconstruct the complete logits in \eqref{eq:complete_frequency_logits}, and perform the RFIS and RPD interventions separately for every layer, Query head, and frequency. The visible-key set is unchanged by either intervention. Under GQA, Query heads that share a KV head are still measured independently because they retain distinct queries, frequency contributions, and complete attention distributions. RFIS and RPD use the same samples and query positions, and all analyses use their raw bounded scores without head-wise normalization.

The RFIS/RPD heatmaps in Figure~\ref{fig:controlled_boundary} and this appendix use panel-specific zero-to-maximum scales. For display, normalized scores $t=s/s_{\max}$ are mapped to $t^{0.20}$ to resolve variation near zero. The colorbars label the corresponding raw scores.

\subsection{Layer-Mean Frequency Structure in Qwen3-4B}
\label{app:qwen_layermean}

\begin{figure}[htbp]
    \centering
    \includegraphics[width=\linewidth]{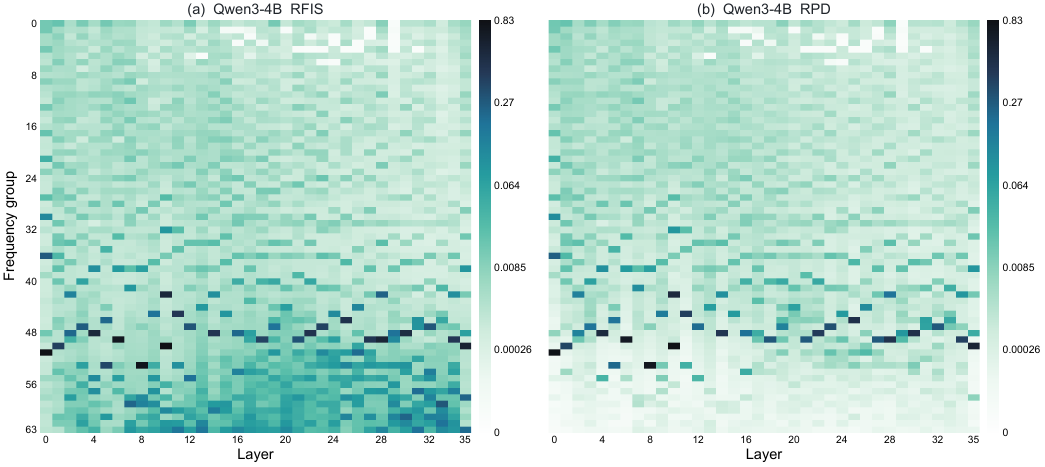}
    \caption{\textbf{Layer-mean RFIS and RPD in Qwen3-4B.} RFIS (left) and RPD (right) are averaged over Query heads in each layer. Frequencies descend from high to low, and each panel uses its own zero-to-maximum scale.}
    \label{fig:qwen3_4b_layermean}
\end{figure}

Figure~\ref{fig:qwen3_4b_layermean} shows multiple bands whose positions vary with depth. Each controlled setting in Figure~\ref{fig:controlled_boundary} shows a narrower dominant band. Qwen3's final pretraining stage increases the sequence length from $4{,}096$ to $32{,}768$~\citep{yangQwen3TechnicalReport2025}. The public training history leaves the effective training length of each layer open, so we describe the observed bands as GPBand-like.

\subsection{Cross-Family Mechanism-Level Consistency}

\begin{figure}[!ht]
    \centering
    \includegraphics[width=\linewidth]{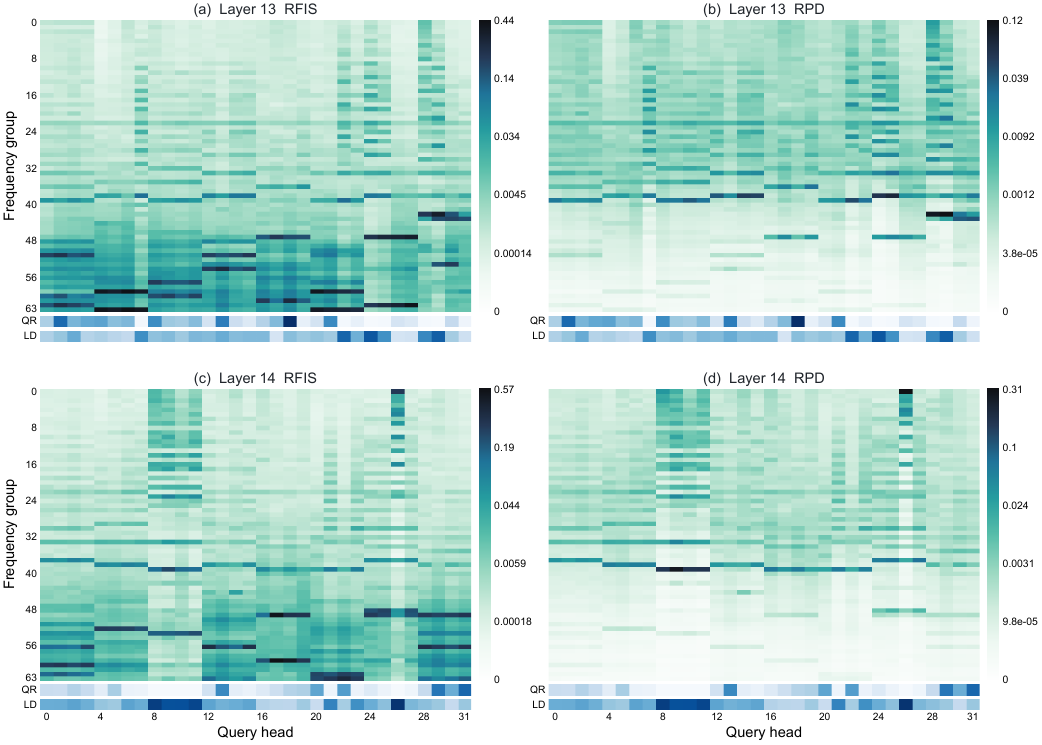}
    \caption{\textbf{Cross-family RoPE-frequency dependence in Llama3.1-8B-Instruct.} Rows show layers $13$ and $14$; columns show RFIS and RPD. Aligned QR and LD strips report retrieval and local-streaming tendencies. Frequencies descend from high to low, and each RFIS/RPD panel uses its own raw zero-to-maximum scale.}
    \label{fig:llama_rfis_rpd_qrld}
\end{figure}

In Llama3.1-8B-Instruct (Figure~\ref{fig:llama_rfis_rpd_qrld}), some heads emphasize slow frequencies with weak RPD, while others show coincident RFIS and RPD near and above a salient intermediate band. Both signatures can coexist within individual heads, as in Qwen3-4B (Figure~\ref{fig:qwen3_4b_rfis_rpd_qrld}).

\FloatBarrier

\subsection{Controlled Analysis of GPBand}
\label{app:controlled_boundary}

\paragraph{Controlled settings.}
The three QKNorm settings in Figure~\ref{fig:controlled_boundary} share an architecture with head dimension $d_h=64$, along with the tokenizer, data, and optimization settings. The $(\theta,L_{\mathrm{train}})=(10^4,2048)$ setting uses the full-RoPE checkpoint in Table~\ref{tab:gpband_pretraining}.

With $F=d_h/2=32$ two-dimensional frequency groups, group $r$ has angular frequency and rotation period
\begin{equation}
    \omega_r=\theta^{-r/F},
    \qquad
    \lambda_r=\frac{2\pi}{\omega_r}=2\pi\theta^{r/F}.
    \label{eq:gpband_period}
\end{equation}
Table~\ref{tab:controlled_gpband_periods} reports the observed periods relative to the training length.

\begin{table}[htbp]
    \centering
    \small
    \caption{\textbf{Rotation periods at the controlled GPBands.}}
    \label{tab:controlled_gpband_periods}
    \begin{tabular}{ccccc}
        \toprule
        $\theta$ & $L_{\mathrm{train}}$ & Group $r$ & Period $\lambda_r$ & $\lambda_r/L_{\mathrm{train}}$ \\
        \midrule
        $10^{4}$ & $2048$ & $21$ & $2650$ & $1.29$ \\
        $10^{4}$ & $2048$ & $22$ & $3533$ & $1.73$ \\
        $10^{6}$ & $2048$ & $14$ & $2650$ & $1.29$ \\
        $10^{6}$ & $2048$ & $15$ & $4080$ & $1.99$ \\
        $10^{4}$ & $4096$ & $24$ & $6283$ & $1.53$ \\
        \bottomrule
    \end{tabular}
\end{table}

\paragraph{Connection to Frequency Bands.}
\citet{okaFREQUENCYBANDSROPE2026} locate a high-norm band from two-dimensional Q/K norms and explain it through the positional variation available to RoPE over the training length. Their appendix jointly analyzes the sine and cosine coordinates through the full covariance
\begin{equation}
    \boldsymbol{\Sigma}(x)
    =
    \operatorname{Cov}_{m\sim\operatorname{Unif}[0,L_{\mathrm{train}}]}
    \!\left[
    \begin{pmatrix}
        \cos(m\omega) \\
        \sin(m\omega)
    \end{pmatrix}
    \right]
    =
    \begin{pmatrix}
        \Sigma_{11}(x) & \Sigma_{12}(x) \\
        \Sigma_{12}(x) & \Sigma_{22}(x)
    \end{pmatrix},
    \quad x=\omega L_{\mathrm{train}},
    \label{eq:frequency_bands_covariance}
\end{equation}
where
\begin{equation}
\begin{aligned}
    \Sigma_{11}(x)
    &=\frac12+\frac{\sin(2x)}{4x}-\left(\frac{\sin x}{x}\right)^2, \\
    \Sigma_{22}(x)
    &=\frac12-\frac{\sin(2x)}{4x}-\left(\frac{1-\cos x}{x}\right)^2, \\
    \Sigma_{12}(x)
    &=\frac{1-\cos(2x)}{4x}
      -\frac{\sin x}{x}\frac{1-\cos x}{x}.
\end{aligned}
\label{eq:frequency_bands_covariance_entries}
\end{equation}
Under a fixed coefficient-norm budget, the maximum centered variation is determined by $\lambda_{\max}(\boldsymbol{\Sigma}(x))$. Maximizing this quantity gives $x^\star\approx4.493409$ and hence
\begin{equation}
    r^\star
    =
    F\log_{\theta}\!\left(\frac{L_{\mathrm{train}}}{x^\star}\right),
    \qquad
    \lambda^\star
    =
    \frac{2\pi L_{\mathrm{train}}}{x^\star}
    \approx1.398L_{\mathrm{train}}.
    \label{eq:frequency_bands_predictor}
\end{equation}
For $L_{\mathrm{train}}=2048$, this predicts $r^\star=21.27$ at $\theta=10^4$ and $r^\star=14.18$ at $\theta=10^6$, with period $2864$ in both cases, matching groups $21$--$22$ and $14$--$15$. For $L_{\mathrm{train}}=4096$ and $\theta=10^4$, it predicts $r^\star=23.68$ and period $5727$, close to observed group $24$ with period $6283$. In layer $0$ of the length-$4096$ model, RFIS peaks at the lowest-frequency group, while RPD shows a weak peak at group $24$.

\begin{figure}[htbp]
    \centering
    \includegraphics[width=\linewidth]{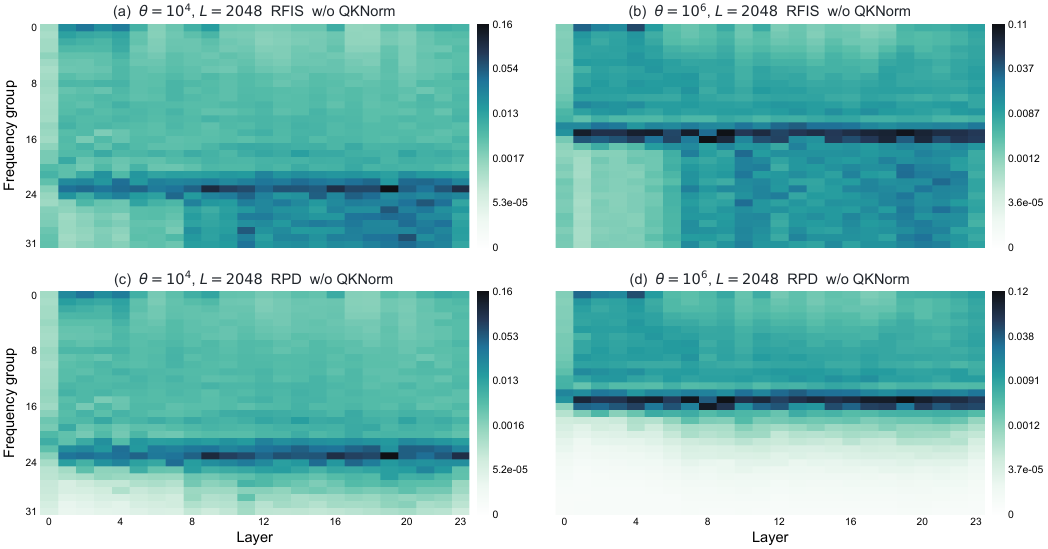}
    \caption{\textbf{Controlled models without QKNorm.} Rows show layer-mean RFIS and RPD at training length $2048$; columns show $\theta=10^4$ and $10^6$. Relative to Figure~\ref{fig:controlled_boundary}, the bands shift toward lower frequencies.}
    \label{fig:controlled_no_qknorm}
\end{figure}

\paragraph{Effect of QKNorm.}
The norm-defined empirical band in \citet{okaFREQUENCYBANDSROPE2026} differs from the theoretical index by an approximately $1.1$ factor in several pretrained models. Figure~\ref{fig:controlled_no_qknorm} shows that removing QKNorm shifts the $\theta=10^4$ band from groups $21$--$22$ toward group $23$ and the $\theta=10^6$ band from groups $14$--$15$ toward group $15$--$16$, after layer $0$.

\clearpage

\begin{figure}[!ht]
    \centering
    \includegraphics[width=\linewidth]{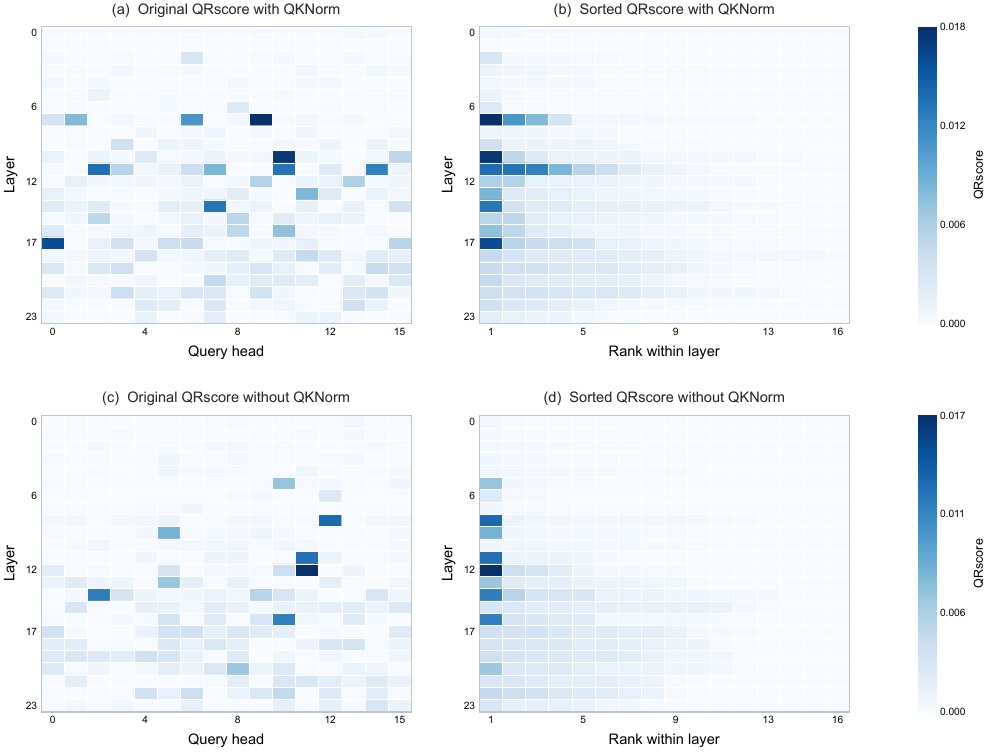}
    \caption{\textbf{Retrieval tendency across layers in our $380$M Transformers trained at $2$K.} Rows show the models with QKNorm (top) and without QKNorm (bottom). Columns show head-wise QRscores in their original order and sorted from high to low within each layer. Both panels in each row share a raw zero-to-maximum color scale, with negative scores displayed at zero.}
    \label{fig:controlled_qr_depth}
\end{figure}

\begin{figure}[!ht]
    \centering
    \includegraphics[width=\linewidth]{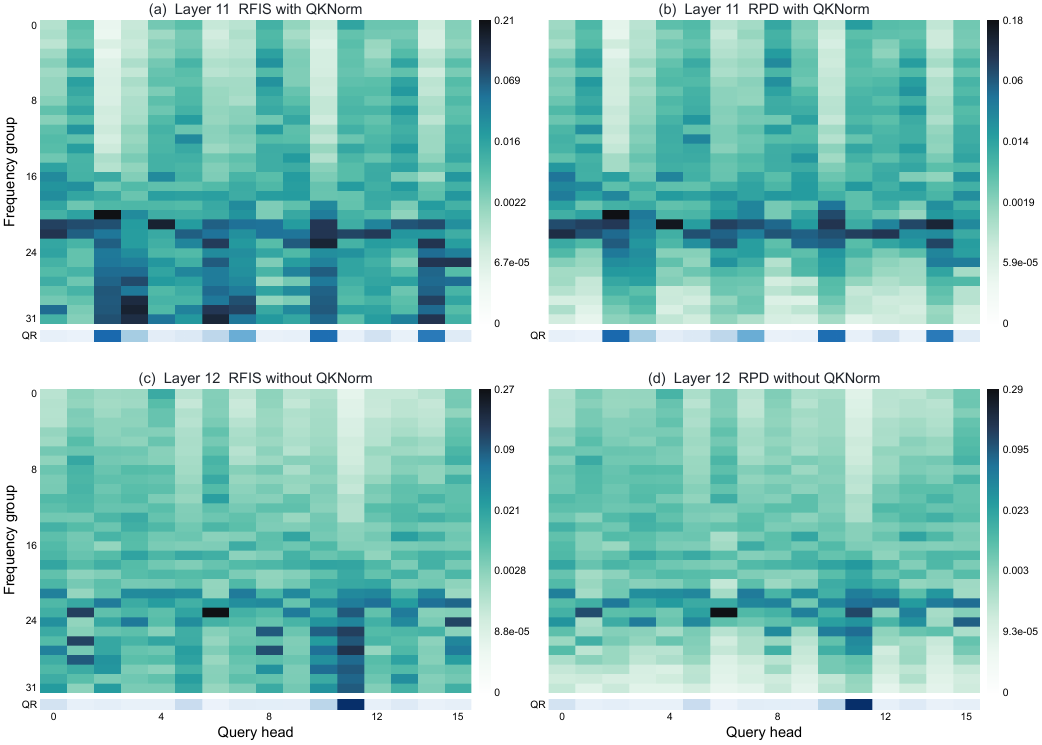}
    \caption{\textbf{Implicit functional differentiation in our $380$M Transformers trained at $2$K.} RFIS (left) and RPD (right) in Layer $11$ of the full-RoPE model with QKNorm in Table~\ref{tab:gpband_pretraining} (top) and Layer $12$ of the original Transformer without QKNorm (bottom). Layer indices are zero-based. QRscore strips are aligned below the frequency heatmaps for the same Query heads and normalized within each checkpoint for display. Frequencies descend from high to low. Each frequency heatmap uses its own raw zero-to-maximum scale.}
    \label{fig:controlled_head_frequency}
\end{figure}

\paragraph{Functional differentiation in our pretrained Transformers.}
\label{par:pretrained_functional_differentiation}
Figure~\ref{fig:controlled_qr_depth} shows the original and within-layer sorted QRscores for our $380$M Transformers with and without QKNorm, both trained at $2$K. High retrieval scores concentrate at intermediate depths in both models, consistent with the layer-specific pattern in Qwen3-4B. Figure~\ref{fig:controlled_head_frequency} shows how these tendencies relate to RFIS and RPD in Layer $11$ with QKNorm and Layer $12$ without QKNorm. In both models, several heads with high QRscore rely on low frequencies with substantial RFIS and relatively weak RPD. Other heads emphasize the salient band and can exhibit isolated RFIS peaks at lower frequencies. The recurrence of these patterns in smaller pretrained Transformers supports testing the design principles at this scale.

\clearpage

\FloatBarrier

\subsection{Length Extrapolation of the GPBand Pretraining Models}
\label{app:gpband_extrapolation}

\begin{figure}[!ht]
    \centering
    \includegraphics[width=\linewidth]{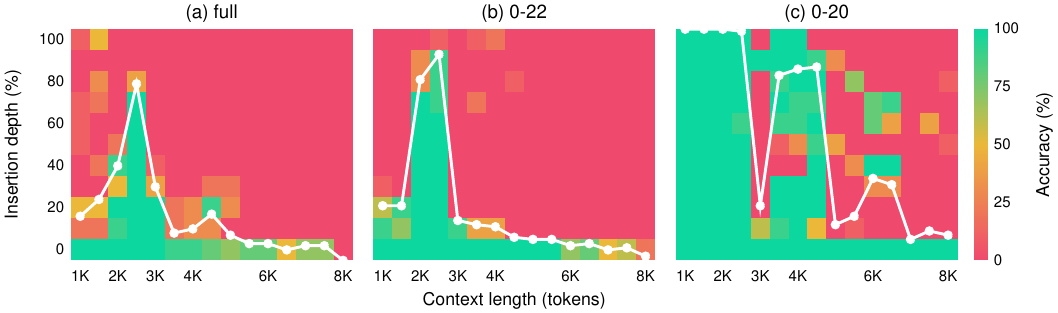}
    \caption{\textbf{NIAH-Single-2 heatmaps for the GPBand pretraining models.} Panels follow the order of Table~\ref{tab:gpband_pretraining}: full RoPE, groups $0$--$22$, and groups $0$--$20$. Axes, insertion depths, and white depth-averaged accuracy curves follow Figure~\ref{fig:niah_all_380m}. The $0$--$20$ model has a higher grid mean but exhibits pronounced nonmonotonic failures, including at the $3$K context budget.}
    \label{fig:gpband_pretraining_niah}
\end{figure}

Figure~\ref{fig:gpband_pretraining_niah} evaluates the three checkpoints in Table~\ref{tab:gpband_pretraining} using the same dense NIAH-Single-2 protocol as Figure~\ref{fig:niah_all_380m}: $15$ context budgets from $1$K to $8$K in $512$-token steps, $11$ insertion depths, and $10$ examples per cell. Appendix~\ref{app:experimental_details} describes this protocol, whose sampling differs from the task-level evaluation in Table~\ref{tab:gpband_pretraining}.

The $0$--$20$ model improves the full-grid mean to $53.64\%$, compared with $18.85\%$ for full RoPE and $21.03\%$ for $0$--$22$, but does not extrapolate consistently. Its depth-averaged accuracy falls from $99.09\%$ at budget $2560$ to $23.64\%$ at $3072$, then recovers to $80.00\%$ at $3584$ and $82.73\%$ at $4096$. It falls again to $15.45\%$ at $5120$ and reaches only $10.91\%$ at $8192$. The strong $4$K result in Table~\ref{tab:gpband_pretraining} therefore captures a relative improvement at a selected length, not stable performance across extrapolation lengths.

Withholding GPBand leaves the position-sensitive mechanism free to access global context. In the $0$--$20$ model, even the slowest retained rotation has a period of approximately $1987$ tokens, shorter than the $2$K training length. Although each retained frequency wraps within that span, combinations of frequencies can still distinguish global offsets. The model may therefore learn residual global positional fitting through those combinations, offering a possible explanation for its remaining extrapolation failures. The heatmaps establish the instability, rather than identifying the combinations responsible for any particular dip. Together with the comparable commonsense reasoning and real-world retrieval performance relative to $0$--$22$ in Table~\ref{tab:gpband_pretraining}, these results show that improved but unstable extrapolation can emerge without a broad loss of in-window capability. Restricting frequencies consequently does not replace the local positional modeling and separate position-independent global retrieval called for by Principle~2.

\FloatBarrier

\subsection{Post-training RoPE masking}
\label{app:gpband_inference_ablation}

We replace the RoPE rotations for two adjacent frequency groups at a time with the identity in the full-RoPE $380$M QKNorm checkpoint from Table~\ref{tab:gpband_pretraining}, trained at $2$K, preserving the channels, learned weights, and causal mask. Table~\ref{tab:gpband_mask_ablation} reports the unmasked checkpoint and all ten masks, using zero-based indices. Groups $21$--$22$ form the GPBand. The restricted-RoPE variants in Table~\ref{tab:gpband_pretraining} are separately pretrained.

The GPBand mask produces the lowest six-task retrieval average, reducing it from $27.72\%$ to $21.57\%$, while commonsense accuracy remains nearly unchanged ($42.66\%$ versus $42.63\%$). FDA, SWDE, and NQ, highlighted in the main text, are the only tasks containing original prompts longer than $2048$ tokens; their longer examples are cropped to fit the $2$K evaluation budget. Only a subset of SWDE examples is this long. SQuAD, TriviaQA, and DROP are all shorter and require no cropping. The three highlighted tasks lose more than $5$ points each, whereas the other three lose less than $2$ points each. Table~\ref{tab:gpband_mask_ablation} reports all six task scores. Appendix~\ref{app:retrieval_context} specifies the length statistics and the reserved generation budget: actual prompts contain at most $2000$ tokens.

These retrieval losses show that the model relies on GPBand rotations for useful positional computation. Combined with RFIS/RPD and the shift of GPBand with training length, they support positional fitting to the training span. Post-training masking tests this learned reliance; the separately pretrained variants in Section~\ref{subsec:controlled_boundary} test how the availability of these rotations during learning affects extrapolation.

\begin{table}[!ht]
\centering
\footnotesize
\setlength{\tabcolsep}{3pt}
\caption{\textbf{Masking adjacent RoPE groups after training.} Scores are percentages. Common. is the commonsense mean; Retr. is the mean across all six cloze tasks shown. $\dagger$ marks the only tasks containing original prompts longer than $2048$ tokens (Table~\ref{tab:retrieval_contexts}); longer examples are cropped to fit the $2$K budget. The unmasked checkpoint is the full-RoPE model in Table~\ref{tab:gpband_pretraining}. Bold marks the GPBand intervention.}
\label{tab:gpband_mask_ablation}
\begin{tabular*}{\linewidth}{@{\extracolsep{\fill}}lrrrrrrrr@{}}
\toprule
Mask & Common. & Retr. & FDA$^{\dagger}$ & SWDE$^{\dagger}$ & NQ$^{\dagger}$ & SQuAD & TriviaQA & DROP \\
\midrule
None & 42.66 & 27.72 & 17.60 & 29.34 & 14.67 & 34.65 & 49.53 & 20.51 \\
1--2 & 42.48 & 25.43 & 14.79 & 25.29 & 12.73 & 32.17 & 47.87 & 19.74 \\
5--6 & 42.49 & 25.92 & 12.43 & 28.17 & 13.40 & 32.94 & 48.28 & 20.32 \\
9--10 & 42.25 & 25.28 & 12.79 & 28.53 & 14.44 & 30.16 & 46.45 & 19.31 \\
13--14 & 42.12 & 23.81 & 12.79 & 27.00 & 13.43 & 25.87 & 45.62 & 18.16 \\
17--18 & 42.45 & 23.44 & 10.25 & 25.74 & 12.80 & 28.02 & 46.56 & 17.25 \\
19--20 & 42.81 & 27.35 & 13.97 & 29.25 & 13.62 & 35.82 & 50.12 & 21.32 \\
\textbf{21--22} & \textbf{42.63} & \textbf{21.57} & \textbf{2.18} & \textbf{17.64} & \textbf{8.77} & \textbf{33.65} & \textbf{48.46} & \textbf{18.74} \\
23--24 & 42.74 & 24.87 & 12.79 & 23.04 & 12.48 & 33.04 & 48.82 & 19.02 \\
25--26 & 42.73 & 26.42 & 16.42 & 26.19 & 13.97 & 33.61 & 48.82 & 19.50 \\
29--30 & 42.76 & 27.64 & 18.06 & 29.07 & 14.70 & 34.55 & 49.35 & 20.12 \\
\bottomrule
\end{tabular*}
\end{table}

\FloatBarrier

\subsection{Gemma-7B and the Limits of Norm-Based Analysis}

\begin{figure}[htbp]
    \centering
    \includegraphics[width=\linewidth]{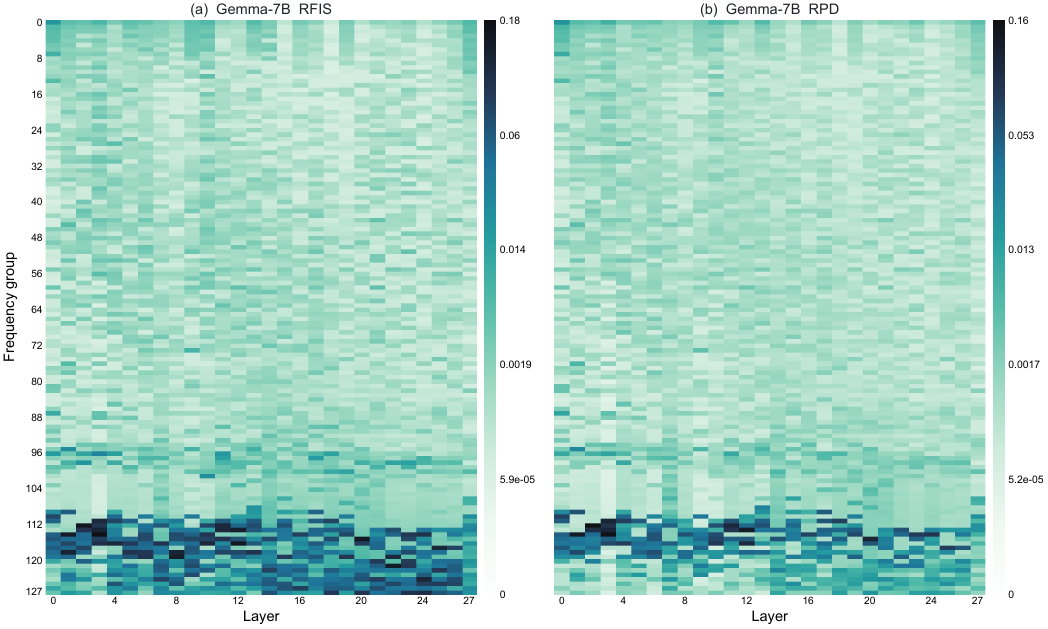}
    \caption{\textbf{Layer-mean RFIS and RPD in Gemma-7B.} The dominant position-dependent structure lies near groups $110$--$120$, below the training-length prediction near group $104$; a weaker band appears near group $96$. Panels use independent raw zero-to-maximum scales.}
    \label{fig:gemma_layermean}
\end{figure}

Gemma-7B provides a useful comparison with the norm-based analysis of \citet{barberoWeGoWhat2024}. The RFIS/RPD detections in Figure~\ref{fig:gemma_layermean} show that its dominant position-dependent Band extends unusually far into the low-frequency range. With $F=128$, $\theta=10^4$, and an approximately $8000$-token training length, \eqref{eq:frequency_bands_predictor} predicts $r^\star\approx104.0$ and $\lambda^\star\approx11{,}186$. Groups $100$ and $110$ have periods $8379$ and $17{,}206$, respectively, so the expected training-length-related band lies between them. RPD, however, remains strong through approximately group $120$, whose period is $35{,}333$. This extension may reflect the absence of QKNorm or an undocumented training pipeline. A weaker band near group $96$, with period $6283<8000$, is also visible, but its role remains unclear.

This result changes the interpretation of the low-frequency dependence reported by \citet{barberoWeGoWhat2024}: a substantial portion of the region identified through Q/K norms is highly position dependent rather than position-invariant retrieval structure. Frequency location and norm magnitude therefore cannot directly determine whether a RoPE channel depends on positional modulation; RPD is required for that distinction.

\begin{figure}[!ht]
    \centering
    \includegraphics[width=0.88\linewidth]{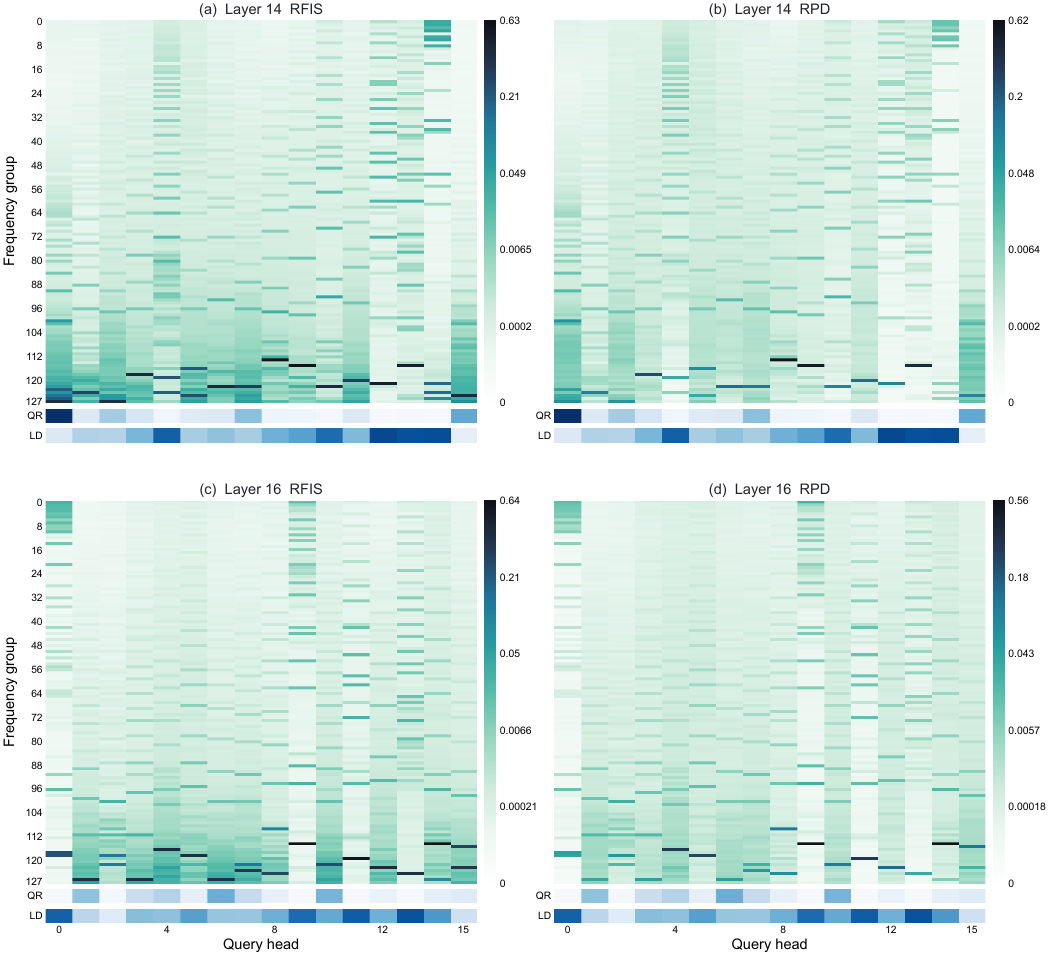}
    \caption{\textbf{Head-level RoPE-frequency dependence in Gemma-7B.} Rows show layers $14$ and $16$; columns show RFIS and RPD with aligned QR and LD strips. Frequencies descend from high to low, and each RFIS/RPD panel uses its own raw zero-to-maximum scale.}
    \label{fig:gemma_rfis_rpd_qrld}
\end{figure}

In Gemma-7B layers $14$ and $16$ (Figure~\ref{fig:gemma_rfis_rpd_qrld}), heads with high QRscore often emphasize low frequencies with comparatively weak RPD. Other heads exhibit coincident RFIS and RPD near or above the shifted band.

\FloatBarrier

\section{Theoretical Details of RFIS and RPD}
\label{app:theory}

\paragraph{Relation to norm-based analysis.}
\citet{barberoWeGoWhat2024} approximate RoPE-frequency usage using mean per-frequency Q/K norms, motivated by the fact that the Cauchy--Schwarz inequality bounds the magnitude of each frequency's contribution in terms of these norms. Because norms discard directional alignment between Q and K, this magnitude-only approximation introduces error when estimating each frequency's realized contribution to attention. RFIS instead removes the exact contribution of one frequency from the Q/K attention computation while holding all others fixed and measures the resulting change in the complete attention distribution, providing a more faithful frequency-importance metric. RPD then applies a further controlled intervention: it leaves the frequency's Q/K vectors unchanged and replaces only its relative RoPE rotation with the identity, thereby isolating dependence on rotary positional modulation, which norm-based analysis cannot measure.

\subsection{Boundedness of the normalized JS readout}

Throughout the RFIS and RPD definitions, $\log$ denotes the natural logarithm. Let $\mathbf m=(\mathbf p+\mathbf q)/2$. Non-negativity follows from the two KL terms in \eqref{eq:js_definition}. For the upper bound, $m_j\geq p_j/2$, so for every $p_j>0$,
\begin{equation}
\log\frac{p_j}{m_j}\leq\ln2.
\end{equation}
Therefore,
\begin{equation}
D_{\mathrm{KL}}(\mathbf p\Vert\mathbf m)
=\sum_jp_j\log\frac{p_j}{m_j}
\leq\ln2.
\end{equation}
The same argument applies to $\mathbf q$, giving
\begin{equation}
0\leq D_{\mathrm{JS}}(\mathbf p,\mathbf q)\leq\ln2.
\label{eq:appendix_js_bound}
\end{equation}
Equality at the upper bound requires disjoint supports. A softmax with finite valid logits assigns positive probability throughout the common support, so attention distributions generally approach rather than attain the normalized value $1$.

\section{Additional Experimental Details}
\label{app:experimental_details}

\paragraph{Pretraining.}
For the architecture comparisons in Section~\ref{sec:experiments}, we train from scratch on the \texttt{sample-100BT} subset of FineWeb-Edu~\citep{penedoFineWebDatasetsDecanting2024} with the Mistral tokenizer (vocabulary size $32{,}000$) and sequences of $2{,}048$ tokens. The approximately $380$M and $1.4$B models use global batch sizes of $256$ and $1{,}024$, respectively, with mixed precision. Within a scale, the models use the same data order, optimizer schedule, and evaluation protocols. The total Q/K/V width allocated to the sequence mixers in a layer is $1{,}024$ and $2{,}048$, respectively. At $380$M, NoPE FA heads use Q/K/V dimension $64$ and GDN heads use $128$. At $1.4$B, both use $128$. Thus, the width allocated to a component and its fraction of heads can differ. The architectures have similar parameter counts within each scale and differ in sequence-mixer parameters and operations. All models use fused AdamW with $\beta_1=0.9$, $\beta_2=0.95$, $\epsilon=10^{-15}$, weight decay $0.1$, and a peak learning rate of $3\times10^{-4}$. We apply $1{,}024$ warmup steps followed by cosine decay to $10\%$ of the peak rate and clip the gradient norm at $1.0$.

\paragraph{Evaluation.}
We evaluate with the LM Evaluation Harness~\citep{sutawikaEleutherAILmevaluationharnessMajor2023}. Perplexity uses WikiText~\citep{merityPointerSentinelMixture2017} and the OpenAI-formatted LAMBADA benchmark~\citep{papernoLAMBADADatasetWord2016}. Commonsense reasoning uses LAMBADA, HellaSwag~\citep{zellersHellaSwagCanMachine2019}, PIQA~\citep{biskPIQAReasoningPhysical2020}, ARC-E and ARC-C~\citep{clarkThinkYouHave2018}, WinoGrande~\citep{sakaguchiWinoGrandeAdversarialWinograd2019}, and OpenBookQA~\citep{mihaylovCanSuitArmor2018}.

\paragraph{Real-world retrieval context.}
\label{app:retrieval_context}
The cloze-completion suite~\citep{aroraSimpleLinearAttention2024,aroraJustReadTwice2024} covers FDA and SWDE, SQuAD~\citep{rajpurkarKnowWhatYou2018a}, Natural Questions (NQ)~\citep{kwiatkowskiNaturalQuestionsBenchmark2019}, TriviaQA~\citep{joshiTriviaQALargeScale2017}, and DROP~\citep{duaDROPReadingComprehension2019}. It uses a $2{,}048$-token context budget. The harness reserves $48$ tokens for generation and up to $48$ tokens for the question suffix and tokenization boundaries. Cleaned contexts longer than $1{,}952$ tokens are cropped around the answer, and prompts use at most $2{,}000$ tokens. Shorter contexts retain their original length. Table~\ref{tab:retrieval_contexts} reports prompt lengths before truncation using the evaluation tokenizer.

\begin{table}[htbp]
\centering
\small
\caption{\textbf{Validation prompts in the cloze retrieval suite.} Lengths are measured before cropping with the evaluation tokenizer. \emph{Cropped} counts contexts exceeding the $1{,}952$-token threshold.}
\label{tab:retrieval_contexts}
\begin{tabular}{lrrrrr}
\toprule
Task & Samples & Cropped & Median tokens & $90$th percentile & Maximum \\
\midrule
FDA      & 1,102 & 954   & 2,373 & 2,506 & 2,738 \\
SWDE     & 1,111 & 73    & 1,230 & 1,900 & 2,179 \\
SQuAD    & 2,984 & 0     & 169   & 298   & 853 \\
NQ       & 3,157 & 3,091 & 3,312 & 4,536 & 10,849 \\
TriviaQA & 1,688 & 0     & 337   & 385   & 491 \\
DROP     & 2,087 & 0     & 283   & 451   & 1,560 \\
\bottomrule
\end{tabular}
\end{table}

\paragraph{Synthetic retrieval and length extrapolation.}
We use the RULER NIAH tasks NIAH-Single-1, NIAH-Single-2, and NIAH-Single-3~\citep{hsiehRULERWhatsReal2024}. At $380$M, the fixed-depth NIAH-Single-2 evaluation follows the length--depth protocol of \citet{gkamradtGkamradtLLMTest_NeedleInAHaystack2026} with the Paul Graham Essays haystack and a $128$-token generation budget. The grid uses $15$ context budgets ($1{,}024,1{,}536,\ldots,8{,}192$) and $11$ insertion depths ($0\%,10\%,\ldots,100\%$), with $10$ examples per cell and $1{,}650$ per model. The grid mean weights all cells equally. The dense grid and the task-level RULER NIAH evaluations use different sampling schemes.

For the $1.4$B comparison in Figure~\ref{fig:niah_extrapolation_1p4b}, we use context budgets $1{,}024$, $2{,}048$, $4{,}096$, $8{,}192$, $16{,}384$, and $32{,}768$, with the same $11$ insertion depths. All three checkpoints use SWAN-style logarithmic query scaling~\citep{puvvadaSWANEfficientScalable2025a} with $\alpha=300$ in their NoPE FA components at inference, without additional training. These scaled dense-grid results are reported separately from the task-level scores in Table~\ref{tab:main_retrieval}.

\paragraph{Short-length instability at small scale.}
\label{app:dense_niah_robustness}
The $380$M Transformer and HwH-std-r exhibit anomalously poor NIAH-Single-2 performance at $1$K. Rather than returning the seven-digit target, their specific erroneous responses include \texttt{\_saltimbocca\_}, \texttt{\_saltintesta\_}, and the underscore-only blank \texttt{\_\_\_\_\_\_\_\_}. These are not simply repeated phrases that dominate the haystack: for example, \texttt{\_saltimbocca\_} occurs only once in the corresponding sample, roughly $10$--$20\%$ into the sequence, yet is selected as the answer. This behavior suggests that the affected small-scale architectures can lock onto a salient distractor or fall into a blank-response mode even within short contexts, reflecting limited short-sequence robustness rather than length-extrapolation behavior. The $1.4$B Transformer remains stable at both $1$K and $2$K, further showing that the phenomenon is scale- and configuration-dependent.

\paragraph{Controlled GPBand models.}
The controlled RoPE Transformers use the tokenizer, model scale, and optimization settings of the $380$M experiments. Appendix~\ref{app:controlled_boundary} gives the band-location settings.

\section{Allocation and component configurations}
\label{app:additional_ablation}

Table~\ref{tab:appendix_hwh_variants} records the HwH component and width allocations. Outer layers are the shallow and deep layers outside the middle half, excluding the first and last layers. The $1{:}3$ middle-layer width ratio parallels Inter's one-in-four full-attention allocation, and the $1{:}7$ outer-layer ratio halves the global width fraction from one quarter to one eighth.

\begin{table}[htbp]
\centering
\caption{Configurations of the HwH ablation models. Ratios denote global-component:local-component Q/K/V dimensions.}
\label{tab:appendix_hwh_variants}
\small
\begin{tabular*}{\linewidth}{@{\extracolsep{\fill}}llllll}
\toprule
Model & Global & Local & First/last & Middle half & Outer layers \\
\midrule
HwH-uni   & NoPE FA & GDN & $1{:}3$ & $1{:}3$ & $1{:}3$ \\
HwH-rmfl  & NoPE FA & GDN & all local & $1{:}3$ & $1{:}3$ \\
HwH-std   & NoPE FA & GDN & all local & $1{:}3$ & $1{:}7$ \\
HwH-swa   & NoPE FA & SWA & all local & $1{:}3$ & $1{:}7$ \\
HwH-std-r & RoPE FA & GDN & all local & $1{:}3$ & $1{:}7$ \\
\bottomrule
\end{tabular*}
\end{table}

\subsection{Detailed results at 380M}
\label{app:detailed_ablation_results}

Tables~\ref{tab:component_lm_reasoning} and~\ref{tab:component_retrieval} provide the task-level results for the models in Table~\ref{tab:principle_ablation}.

\begin{table}[htbp]
\centering
\caption{\textbf{Language modeling and commonsense reasoning at $380$M.} The first two columns report word perplexity (lower is better). All remaining columns report accuracy (\%). Avg. is the mean of the seven task accuracies.}
\label{tab:component_lm_reasoning}
\label{tab:allocation_lm_reasoning}
\scriptsize
\setlength{\tabcolsep}{3.1pt}
\begin{tabular*}{\linewidth}{@{\extracolsep{\fill}}lrrrrrrrrrr}
\toprule
Model & WikiText $\downarrow$ & LAMBADA $\downarrow$ & LAMBADA & HellaSwag & PIQA & ARC-E & ARC-C & WinoGrande & OpenBookQA & Avg. \\
\midrule
Transformer & 29.10 & 38.26 & 33.09 & 39.05 & 66.54 & 56.90 & 27.73 & 51.22 & 21.60 & 42.30 \\
GDN & 28.71 & 36.03 & 31.36 & 39.55 & 67.03 & 58.12 & 26.88 & 49.33 & 23.00 & 42.18 \\
Inter & 27.69 & 36.05 & 31.65 & 39.52 & 65.83 & 57.87 & 27.30 & 50.28 & 22.00 & 42.07 \\
HwH-uni & 27.63 & 34.38 & 32.66 & 40.18 & 66.97 & 59.01 & 27.39 & 52.88 & 21.40 & 42.93 \\
HwH-rmfl & 27.94 & 33.85 & 33.55 & 39.63 & 66.76 & 57.79 & 27.65 & 53.35 & 21.40 & 42.88 \\
HwH-std & 27.41 & 34.84 & 32.76 & 39.93 & 67.57 & 57.87 & 28.16 & 50.75 & 22.40 & 42.78 \\
HwH-std-r & 27.32 & 38.15 & 31.01 & 40.01 & 66.87 & 57.62 & 29.52 & 52.64 & 20.80 & 42.64 \\
HwH-swa & 28.97 & 38.54 & 32.76 & 39.11 & 66.81 & 56.19 & 27.22 & 51.85 & 21.00 & 42.13 \\
\bottomrule
\end{tabular*}
\par\medskip
\normalsize
\caption{\textbf{Cloze retrieval at $380$M.} Scores are percentages. Avg. is the mean of the six task scores.}
\label{tab:component_retrieval}
\label{tab:allocation_retrieval}
\small
\begin{tabular*}{\linewidth}{@{\extracolsep{\fill}}lrrrrrrr}
\toprule
Model & FDA & SWDE & SQuAD & NQ & TriviaQA & DROP & Avg. \\
\midrule
Transformer & 14.88 & 24.48 & 31.77 & 13.78 & 46.68 & 18.11 & 24.95 \\
GDN & 6.81 & 11.34 & 26.88 & 11.66 & 46.56 & 16.87 & 20.02 \\
Inter & 29.95 & 27.45 & 33.45 & 16.63 & 49.17 & 17.15 & 28.97 \\
HwH-uni & 21.78 & 27.18 & 31.67 & 15.36 & 48.05 & 18.54 & 27.10 \\
HwH-rmfl & 19.96 & 24.75 & 33.34 & 16.69 & 48.58 & 18.59 & 26.99 \\
HwH-std & 24.14 & 31.05 & 32.74 & 17.29 & 49.17 & 19.45 & 28.98 \\
HwH-std-r & 6.53 & 21.15 & 31.74 & 14.06 & 48.93 & 19.60 & 23.67 \\
HwH-swa & 30.22 & 26.73 & 31.90 & 17.58 & 47.04 & 22.47 & 29.32 \\
\bottomrule
\end{tabular*}
\end{table}

\clearpage

\subsection{SWA replacement in Inter}
\label{app:inter_swa_ablation}

\begin{figure}[!htb]
\centering
\includegraphics[width=\linewidth]{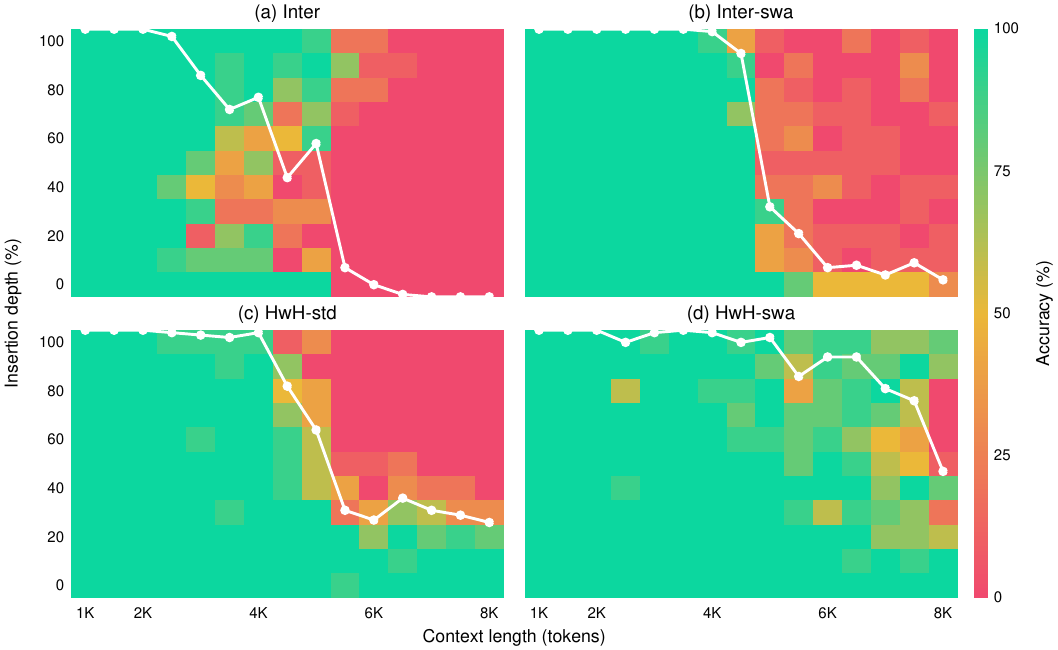}
\caption{\textbf{NIAH-Single-2 heatmaps for the $380$M SWA comparison.} The top row shows layer-wise Inter variants and the bottom row head-wise HwH variants. The left column uses GDN for positional modeling and the right uses $128$-token SWA. Axes and white curves follow Figure~\ref{fig:niah_all_380m}.}
\label{fig:niah_inter_swa_380m}
\end{figure}

Inter-swa retains Inter's layer-wise organization, alternating three $128$-token SWA layers with one NoPE FA layer. Figure~\ref{fig:niah_inter_swa_380m} compares it with Inter, HwH-std, and HwH-swa on the same fixed-depth NIAH-Single-2 examples from $1$K to $8$K.

\clearpage

\section{A Unified View of Functional Differentiation, Positional Modeling, Representational Space, and Hybrid Architectures under the Two Principles}
\label{app:functional_mixing}

A RoPE Transformer learns retrieval and positional modeling within the same Q/K space. Our two principles make different aspects of this implicit division explicit. Principle~1 assigns mechanisms to the two functions and mixes them in layer-specific proportions. Principle~2 gives global access to retrieval and confines positional modeling to local context. Together they connect three architectural choices: how much width each function can use, over what distances it operates, and where it is needed across depth.

\subsection{Representational space: how explicit functional differentiation benefits retrieval}
\label{app:retrieval_space}

RFIS shows that retrieval-oriented heads use frequencies below GPBand even where RPD is weak, so their contribution to matching depends little on rotation. At GPBand and higher frequencies, RFIS and RPD reveal a complementary positional function (Figure~\ref{fig:qwen3_4b_rfis_rpd_qrld}). These two uses compete for width when the same head must learn both through RoPE.

Consider $n$ heads of Q/K/V width $h$ and hidden width $d=nh$. Hold query and key contents $x,y\in\mathbb R^h$ fixed along $h$ independent input directions. Let $W_Q,W_K\in\mathbb R^{h\times h}$ be the query and key projections restricted to these directions. For a query at $t$ and key at $s$, $\mathbf R(u)$ denotes the relative RoPE rotation at offset $u=s-t$. Its $j$th two-dimensional block rotates by $u\omega_j$ (\eqref{eq:rope_logit_decomposition}). Moving the same key to offset $v$ gives $\mathbf R(v)$. The unscaled scores are
\[
 s_u(x,y)=x^\top W_Q^\top\mathbf R(u)W_Ky,\qquad
 s_v(x,y)=x^\top W_Q^\top\mathbf R(v)W_Ky .
\]
To test whether a head can distinguish all $h$ content directions at both offsets, consider any full-rank matching rule $x^\top C y$. For example, $C=I$ gives the ordinary dot product. To realize the same rule for every $x,y$ at both offsets, the projections would have to satisfy
\[
 W_Q^\top\mathbf R(u)W_K=C=W_Q^\top\mathbf R(v)W_K .
\]
Since the rank of a product is at most the rank of each factor, $\operatorname{rank}(C)=h$ forces both $h\times h$ projections to be invertible. Cancelling them would give $\mathbf R(u)=\mathbf R(v)$, a contradiction for distinct rotations. For a NoPE head the positional operator is the identity at every offset, so $W_Q=I$ and $W_K=C$ realize the chosen rule. For a single rotary frequency, \citet[Theorem~6.1]{barberoWeGoWhat2024} also show limits on arbitrarily sharp retrieval across positions in long sequences.

Under the frequency division above, let $a$ of each head's $h$ Q/K coordinates serve positional modeling. Even if a RoPE Transformer specializes $r$ heads in retrieval, at most $r(h-a)$ coordinates remain for approximately position-independent matching. Giving positional modeling its own component makes all $rh$ coordinates of the NoPE retrieval heads available, a gain of \emph{at least $ra$ coordinates} at the same head count and width. \citet{chiang-yogatama-2025-rotary} found that fast-rotating dimensions can contribute little to distant retrieval and suggested omitting RoPE from some heads. RFIS/RPD also identify the positional function those frequencies serve. Explicit separation preserves both functions, consistent with the cloze retrieval gains and competitive general performance of Inter and HwH-std at both scales (Section~\ref{sec:experiments}).

Changing the base illustrates this tradeoff. At fixed training length, a larger base moves GPBand to a smaller group index while its period remains on the training scale (Table~\ref{tab:controlled_gpband_periods}). This leaves fewer coordinate pairs on the positional side and more on the slow-frequency side for matching, even though the accessible prefix is unchanged. In an interleaved RoPE/NoPE model, this could shift some retrieval toward RoPE layers at the expense of positional modeling. This is a representation-level interpretation of the effective attention-range changes that \citet{yang2025rnope} observed when fine-tuning their hybrid. The band shift itself was measured in our controlled pretrained Transformers.

\subsection{Positional scope: local modeling and global retrieval}
\label{app:mixing_position}

Giving each function enough width leaves a second question: which one should reach distant tokens? GPBand shifts with training length. Removing its rotations after training preserves Q/K content but impairs in-window retrieval, showing that the trained model uses positional information fitted to that span (Figure~\ref{fig:controlled_boundary}, Appendix~\ref{app:gpband_inference_ablation}). Withholding the band during pretraining improves extrapolation unevenly while positional modeling retains global access (Appendix~\ref{app:gpband_extrapolation}). Removing one band therefore addresses only part of the learned dependence.

A global positional mechanism encounters offsets or retention lags outside the training range when the prefix grows. Slow RoPE coordinates can reach unseen phases \citep{liu2024ropescale}, and changing the relative offset can reverse the ranking of two keys \citep{du2026rope}. Recurrent retention offers another route: a GDN update can survive across a longer history of gates and state transitions (\eqref{eq:gated_delta_rule}). Long-context state-distribution shifts measured in Mamba-2 illustrate this risk in recurrent models \citep{pmlr-v267-buitrago25a}.

Principle~2 lets the retrieval range grow while keeping local positional offsets within the training range. With fixed inputs inside a window, extending the remote prefix leaves the direct positional computation unchanged. Global NoPE retrieval can still access the new content. \citet{yang2025rnope} demonstrated the benefit of this division in interleaved local RoPE and global NoPE layers. Our architectural controls extend it to head-wise hybrids: restoring global RoPE impairs extrapolation at $380$M, and tightening positional locality improves extrapolation over HwH-std at both scales (Figures~\ref{fig:niah_all_380m} and~\ref{fig:niah_extrapolation_1p4b}).

\subsection{Functional allocation in hybrid architectures}
\label{app:learn_allocation}

Retrieval and positional tendencies coexist within Transformer layers and vary with depth (Appendix~\ref{app:cross_model_consistency}). High QRscores cluster in the middle layers. In the first and last layers, QRscore shows no pronounced retrieval tendency and LDscore shows strong local streaming (Figures~\ref{fig:overview} and~\ref{fig:behavioral_detection}). This observation motivated HwH-std's local-only first and last layers.

Prior findings help explain this pattern. First-layer keys from token embeddings cannot distinguish senses of the same token. Combining neighboring tokens can disambiguate them, consistent with first-layer detokenization \citep{kamoda2025detokenization}. Late layers may combine candidate predictions and calibrate probabilities \citep{lad2025stages}. Final-layer retrieval can affect logits directly, with no subsequent attention layer to integrate the retrieved content.

However, this boundary prior leaves the proportions in the intervening layers open. In a sweep of linear-to-full layer ratios, \citet{wangSystematicAnalysisHybrid2025} found that GatedDeltaNet and HGRN2 reached their highest measured RULER recall at different ratios. A single fixed ratio cannot capture this dependence, motivating an allocation rule that learns the mix for each architecture during pretraining.

One possible implementation fixes the \emph{representation width} at $n$ head slots of $h$ channels per layer. Each eligible slot $i$ has two independently parameterized candidates of width $h$: local positional modeling $P_i$ and global NoPE retrieval $R_i$. Two input-independent scalar logits per slot start at zero, giving each candidate weight $1/2$. The resulting output occupies one slot:
\begin{equation}
 (p_i,r_i)=\operatorname{softmax}(\alpha_i^P,\alpha_i^R),\qquad
 z_i(X)=p_iP_i(X)+r_iR_i(X).
 \label{eq:learn_role_allocation}
\end{equation}
During pretraining, the next-token loss jointly trains the candidates and logits. Gradually increasing the entropy penalty $\lambda\sum_i(-p_i\log p_i-r_i\log r_i)$ encourages one weight to approach one and the other zero. Once all eligible slots have stable, near-binary weights, keep each preferred candidate with unit weight and prune the other. The resulting architecture has learned, deterministic per-layer and overall retrieval shares. Training both branches costs extra compute at fixed output width $nh$. Precedents include differentiable operator selection \citep{liu2019darts} and head or layer selection during model conversion \citep{xiaoDuoAttentionEfficientLongContext2024,lan2026morphing}. 

\paragraph{A unified view.} RFIS and RPD revise earlier accounts of RoPE by tracing positional modeling through the middle frequencies to the training-scale GPBand. This division helps explain the limited width available for position-independent retrieval in RoPE heads. Band interventions link global positional fitting to extrapolation, and from-scratch comparisons support the resulting two principles. Together, these results unify research on Transformer functional differentiation, RoPE's effects on model performance, and hybrid architecture design, bringing a phase of mechanistic discovery to a close. The two principles open a new phase in which the functional demands of autoregressive modeling guide the development of better global retrieval mechanisms, more expressive local positional mechanisms, and new architectures that combine them.

\end{document}